\documentclass{article}

\usepackage[round]{natbib}
\usepackage{arxiv}
\usepackage{url}
\usepackage{microtype}
\usepackage{graphicx}
\usepackage{caption}
\usepackage{subcaption}
\usepackage{booktabs}
\usepackage{multirow}
\usepackage{wrapfig}
\usepackage{nameref}
\usepackage{times}
\usepackage{amsmath}
\usepackage{amssymb}
\usepackage{float}
\usepackage{textcomp}
\usepackage{gensymb}
\usepackage{enumitem}
\usepackage{nameref}
\usepackage{dirtytalk}
\usepackage{bookmark}
\usepackage{dblfloatfix}
\usepackage{appendix}
\usepackage{xcolor}
\usepackage{bm}

\hypersetup{
    colorlinks = true,
    linkcolor = black,
    citecolor = blue,
    urlcolor = black,
}

\definecolor{gred}{rgb}{0.86, 0.27, 0.22}
\definecolor{gyellow}{rgb}{0.97, 0.73, 0}

\hypersetup{
pdftitle={Did I do that? Blame as a means to identify controlled effects in reinforcement learning},
pdfsubject={cs.AI},
pdfauthor={Oriol Corcoll, Raul Vicente},
pdfkeywords={reinforcement learning, causality, unsupervised learning},
}

\title{Did I do that? Blame as a means to identify controlled effects in reinforcement learning}

\author{%
  Oriol Corcoll \footnote{}\\
  Institute of Computer Science\\
  University of Tartu\\
  \texttt{oriol.corcoll.andreu@ut.ee}\\
  \And
  Youssef Mohamed\\
  Institute of Computer Science\\
  University of Tartu\\
  \texttt{youssef.mohamed@ut.ee}\\
  \AND
  Raul Vicente \\
  Institute of Computer Science\\
  University of Tartu\\
  \texttt{raul.vicente.zafra@ut.ee}
}

\date{}
\renewcommand{\headeright}{}
\renewcommand{\undertitle}{}

\begin{document}

\maketitle

\begin{abstract}
Identifying controllable aspects of the environment has proven to be an extraordinary intrinsic motivator to reinforcement learning agents.
Despite repeatedly achieving State-of-the-Art results, this approach has only been studied as a proxy to a reward-based task and has not yet been evaluated on its own.
Current methods are based on action-prediction.
Humans, on the other hand, assign blame to their actions to decide what they controlled.
This work proposes Controlled Effect Network (CEN), an unsupervised method based on counterfactual measures of blame to identify effects on the environment controlled by the agent.
CEN is evaluated in a wide range of environments showing that it can accurately identify controlled effects.
Moreover, we demonstrate CEN's capabilities as intrinsic motivator by integrating it in the state-of-the-art exploration method, achieving substantially better performance than action-prediction models.
\end{abstract}

\section{Introduction}
\label{sec:introduction}
The recent success of reinforcement learning (RL) methods in difficult environments such as Hide \& Seek \citep{baker2019}, StarCraft II \citep{AlphaStar2019}, or Dota2 \citep{openai2019dota} has shown the potential of RL to learn complex behavior.
Unfortunately, these methods also show RL's inefficiency to learn \citep{espeholt2018impala, kapturowski2018, Gulcehre2020}, requiring a vast amount of interactions with the environment before meaningful learning occurs.
Consequently, environments with sparse rewards are known to be extremely difficult making imperative a good exploration strategy.
A popular approach to exploration is to introduce behavioral biases in the form of intrinsic motivators \citep{ChentanezIntrinsic, mohamed2015variational}.
This technique aims to facilitate the learning of task-agnostic behavior by producing dense rewards, driving the agent to discover novel states and by doing so increase the chance of discovering the environment's reward.

Numerous motivators have been developed taking inspiration from humans, e.g. curiosity or control \citep{Bellemare2012InvestigatingCA, pathak2017, Burda2018, Choi2019, badia2020}.
Recent work \citep{Choi2019, song2019megareward, badia2020agent57, badia2020} has achieved State-of-the-Art on the Atari benchmark \citep{bellemare2012ale} by rewarding agents for the discovery of novel ways to control their environment.
A common design principle among these methods is the use of an inverse model to predict the chosen action from two consecutive observations.
The hope is that the latent representation learned encloses aspects of the environment controlled by the agent.

\begin{figure}[t]
    \centering
    \includegraphics[width=0.7\linewidth]{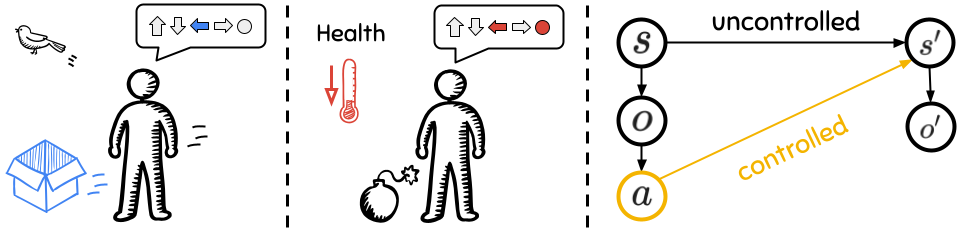}
    \caption{Left) Blame compares an imagined normative world against reality to attribute the movement of the agent and the box to the action. Middle) Using a do-nothing action as normative world is not enough since do-nothing has an effect. Right) Causal graph of a typical RL setting.}
    \label{fig:blame_example}
\end{figure}

A more causal approach is to compare counterfactual worlds \citep{Pearl2009}, i.e., an effect is controllable if the effect would have been different had the agent taken another action. A caveat of this approach is that things become trivially controllable.
Fig. \ref{fig:blame_example} (left) shows an scenario where an agent moves a box by moving left. Here a box becomes controllable even when the agent performs the action "do-nothing" since there is an action, "move-left", that would move the box.
Contrarily, it is believed that humans identify controlled effects by assigning a degree of blame to their actions. In particular, humans compare what happened to a normative world \citep{Halpern2016, morris2018, langenhoff2019, Grinfeld2020}.
If what happened is normal, humans would not assign blame to their actions, e.g. when performing "do-nothing", the box's effect would not be controlled since normally the box would not move. However, it would be considered controlled when performing "move-left" since its normative state is to not move.

This work proposes Controlled Effects Network (CEN), a completely unsupervised approach to identify controlled aspects of the environment based on the human notion of blame.
In contrast to models based on action prediction, CEN is composed of a two-branch forward model that creates a normal and controlled view of the world.
Our experiments show that CEN can disentangle effects precisely, outperforming state-of-the-art approaches to detect controlled effects.
Additionally, we evaluate CEN as an intrinsic motivator by replacing the action-prediction model in Never Give Up's \citep{badia2020} episodic reward with CEN, leading to a substantial gain in performance. 

\section{Identifying controlled effects using blame}
\label{sec:causality}
Our goal is to identify changes in the environment that were controlled by the agent.
This section introduces Individual Causal Effect (ICE), a fundamental measure in causal literature, and frames it in the context of RL.
We show how this measure can be used to identify controllable effects but argue that these are not suitable for RL.
In contrast, the human perception of causality is associated with the concept of blame \citep{gerstenberg2014}.
For example, if lightning hits a forest tree and starts a fire, humans would point to the lightning as the cause of the fire, not the oxygen or wood since they are normally present in the forest.
Consequently, we expand the idea of blame to identify controlled effects by using measures of normality and counterfactuals.

\subsection{Controllable effects}
What does it mean to cause something? \citet{pearl2016} provide an intuitive definition of cause-effect relations: \say{A variable X is a cause of a variable Y if Y, in any way, relies on X for its value}.
This kind of formulation tries to answer questions like \say{does smoking causes cancer?}.
Actual causality, proposed in \citet{Halpern2016}, studies causal relations between individual events of $X$ and $Y$.
It aims to answer questions like, \say{did smoking for 30 years caused David's cancer?}.
In the following, we introduce the concept of causal effect to then define controllable effect in the context of RL.

The individual causal effect (ICE) of an event $X=x$ on a variable $Y_i$ can be measured by comparing counterfactual worlds
\begin{equation}
    \begin{aligned}
        \mathit{ICE}_{Y_i}^{x} \equiv Y^{x}_{i} \neq Y^{\tilde{x}}_{i}\ ,
    \end{aligned}
    \label{eq:ice_generic}
\end{equation}
where $Y^{x}_{i}$ reads as \say{what would the value of an individual $Y_i$ be if $X$ is forced to be $x$}. Similarly, $Y^{\tilde{x}}_i$ describes the value of $Y_i$ when $X$ is forced to not be $x$. Note that the sub-index $i$ refers to an individual, and hence in the following, we use $Y_i^x$ and $Y^x$ interchangeably.

The \textit{fundamental problem of causal inference} states that we can only observe one of these counterfactual worlds and the other needs to be imagined.
Intuitively, Eq. \ref{eq:ice_generic} compares the world where the event $x$ happened to an alternative world where event $x$ had not happened.
Consequently, we say that $x$ has a causal effect on $Y$ if there is an $\tilde{x} \in X$ that satisfies Eq. \ref{eq:ice_generic}.
In the context of RL, $X$ and $Y$ take the form of actions, states and observations. Fig. \ref{fig:blame_example} (right) illustrates the causal relations present in a typical RL setting, where a state $s$ has an effect on both the next state $s'$ and the produced observation $o$ which, in turn, has an effect on the agent's choice of action $a \in \mathcal{A}$. Similarly, an action has an effect on the next state.
Since states are typically not accessible by the agent, we do not use states as variables; nevertheless the same principle can be applied if these are accessible.
We define the perceived effect $e_p^a$ as the difference between consecutive observations when taking action $a$, i.e. $e_p^a \equiv o' - o$. 
As in Eq. \ref{eq:ice_generic}, we say that a perceived effect was controllable by the agent's action when
\begin{equation}
    \begin{aligned}
        \exists{\tilde{a}} \in \mathcal{A} \colon e_p^{a} \neq e_p^{\tilde{a}}\ .
    \end{aligned}
    \label{eq:controllable}
\end{equation}
Since we want to know what elements of the perceived effect are controllable, the inequality is an element-wise operation.
It is important to notice that Eq. \ref{eq:controllable} has far-reaching consequences, for example, an agent next to a box would have a causal effect on it even when not moved since there is a counterfactual world where that box would have moved.
Using Eq. \ref{eq:controllable} as reward, the agent would be rewarded for almost every action at every state!
Note that taking $\tilde{a}$ as a special "do-nothing" action would not work since even doing nothing does something, e.g., Fig. \ref{fig:blame_example} (middle) shows a scenario where doing nothing has an effect on the agent's health. Taking do-nothing as $\tilde{a}$ would not attribute the effect to the agent.
Instead, we would want a more human-like definition of what is controlled where an agent controls a box if moved or its life if a bomb could have been easily avoided.


\subsection{Blame}
It has been shown that human notion of causality is affected by what is normal \citep{kahneman_1986, Cushman2008, Knobe2008, Hitchcock2009}.
Here, we resort to concepts of normality from actual causality to find if the agent's action is to blame for what happened.
\citet{Halpern2014} propose to compare what actually happened with what normally would happen.
Following this idea we use a normative world in replacement to $Y^{\tilde{x}}$
\begin{equation}
    \begin{aligned}
        \mathit{ICE}_Y^x = Y^{x} - \beta_Y\ ,
    \end{aligned}
    \label{eq:ice_normal}
\end{equation}
where $\beta_Y$ is the value $Y$ would normally take.
Such a value is of course contingent to the notion of normality used, which is to us to define.
Note that since we are interested in the magnitude and direction of the effect, Eq. \ref{eq:ice_normal} uses the difference rather than the less specific inequality used in Eq. \ref{eq:ice_generic}.

\textbf{Advantage function as Blame:} A typical use of this formulation is to compute the causal effect of an action on the return $G$ relative to a policy $\pi$ as
\begin{equation}
    \begin{aligned}
        \mathit{ICE}_G^{a}(s) &= G^{a}(s) - \beta_G(s)\\
        &= Q(s,a) - V(s)\\
        &= A(s,a).
    \end{aligned}
    \label{eq:ice_reward}
\end{equation}
$G^{a}(s)$ is the return the agent would get if action $a$ were to be taken at state $s$ and is typically estimated using a state-action value function $Q(s,a)$. The choice of normality for $\beta_G(s)$ is to estimate the expected return with the state-value function $V(s)$, giving us the advantage function $A(s,a)$.

As described in \citet{Sutton2018}, Generalized Value Functions aim to integrate general knowledge of the world; leaving return as special case.
Following the same idea, we can reformulate Eq. \ref{eq:ice_normal} to compute the controlled effect of an action as
\begin{equation}
    \begin{aligned}
        \mathit{ICE}_{e_p}^{a} &= e_p^{a} - \beta_{e_p}\\
        &= e_p^{a} - \mathop{\mathbb{E}}_{\tilde{a}, o'}\Big[e_p^{\tilde{a}}\Big].
    \end{aligned}
    \label{eq:controlled_effects}
\end{equation}
Note that in stochastic environments there may be multiple next observations $o'$ for each action $\tilde{a} \in A$. To simplify notation, the following sections use controlled effect as $e_c^a = \mathit{ICE}_{e_p}^{a}$ and normal effect $e_n = \beta_{e_p}$.
Intuitively, Eq. \ref{eq:controlled_effects} builds a normal world by observing every alternative $e_p$ produced by each action creating an average perceived effect.
Consider the example in Fig. \ref{fig:blame_example} (middle), moving left or doing nothing would make the agent's health decrease. 
Eq. \ref{eq:controlled_effects} would indicate that no changes to the health bar are normal; thus, the loss of health when moving left or staying would be attributed to the agent. On the other hand, moving right would only attribute the change in the agent's position as controlled.
Note that the explosion would never be credited to the agent.

Special care needs to be taken when constructing the normal world $\beta_{e_p}$ for continuous action spaces. Computing counterfactuals on an infinite number of possibilities cannot be done and some approximation needs to be implemented. Although our experiments use discrete actions, the proposed method in the following section is equipped to handle continuous action spaces since it does not compute counterfactuals for each possible action but approximates the normal world directly.
It is also important to notice that the controlled effects Eq. \ref{eq:controlled_effects} can identify in a partially observable setting ($o \neq s$), are constrained to those observed by the agent. Nevertheless, humans cannot perceive every change in state but can identify relevant controlled effects for their survival and joy.

\section{Unsupervised learning of controlled effects}
\label{sec:methods}
In practice, we do not have access to every world and cannot compute Eq. \ref{eq:controlled_effects} directly.
We propose an unsupervised method that disentangles controlled and normal effects only using perceived effects as a self-supervised training signal.

\begin{figure}[t]
    \centering
    \includegraphics[width=0.8\linewidth]{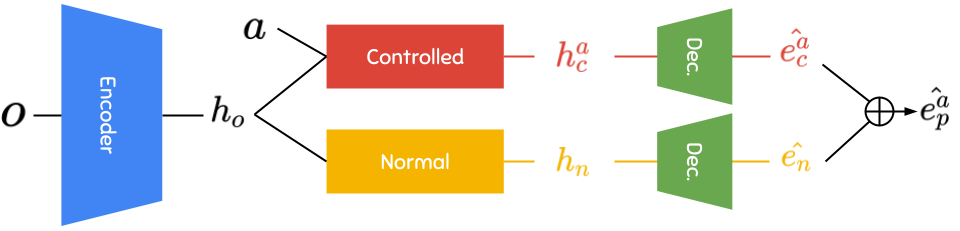}
    \caption{CEN divides the latent space of a forward model into controlled and normal branches. Each branch disentangles controlled and normal effects and decodes each into pixel space independently.}
    \label{fig:model}
\end{figure}

Here, we introduce \textbf{Controlled Effects Network (CEN)}, depicted in Fig. \ref{fig:model}.
CEN is based on a forward model, where observation and action are used to predict the outcome of performing such action on the environment.
In contrast to conventional forward models, CEN divides its latent space into controlled and normal representations; similarly to Dueling Networks \citep{Wang2015}. These two representations approximate the controlled and normal effects in latent space. A decoder converts these latent representations into pixel space allowing to estimate $e^a_c + e_n = e^a_p$ as in Eq. \ref{eq:controlled_effects}.

The \textbf{controlled branch} has privileged access to the action; having only this branch would make CEN a regular forward model, i.e., the controlled branch alone can predict the perceived effect resulting from the action.
Then, why do we need the \textbf{normal branch}? The role of the normal branch is to force the controlled branch to predict only what is not predictable from the observation alone and hence, modeling what is controlled by the agent.
In a way, the normal branch acts as a distillation mechanism where only what can be controlled will be represented by the controlled branch.
To promote the controlled branch to model only controlled effects, we use the following loss
\begin{equation}
    \begin{aligned}
        \mathcal{L} = \text{MSE}\left(\textcolor{gred}{\bm{\hat{e_c^a}}} + \textcolor{gyellow}{\bm{\hat{e_n}}}, e_p^a\right) + \alpha\ \text{MSE}\left(\textcolor{gyellow}{\bm{\hat{e_n}}}, e_p^a\right),
    \end{aligned}
    \label{eq:loss}
\end{equation}
where the first part is the reconstruction loss in which the predicted target $\hat{e_p^a} = \hat{e_c^a} + \hat{e_n}$ is compared to the perceived effects provided by the environment. The second part of the loss enforces the network to use the normal branch as much as possible to model the world. Since this branch cannot predict everything without the action the model will converge to the expected effect due to the MSE loss.
Additionally, a hyperparameter $\alpha$ regulates how much the normal branch should model the environment. In practice, we found that this hyperparameter creates an agreement between branches on uncertain futures which seemed to be critical in environments with stochastic entities.

Let us look again at the example in Fig. \ref{fig:blame_example} (middle) and assume the agent picks the do-nothing action. The normal branch is encouraged to model the bomb since it does not depend on the action. Furthermore, it should not predict any change in health since what is normal is for health to not change. Thus, the controlled branch must model the change in health.

\section{Experiments}
\label{sec:experiments}
This section evaluates CEN\footnote{Networks, training and evaluation have been implemented using PyTorch \citep{pytorch}, NumPy \citep{harris2020array} and PFRL \citep{Fujita2021}; our experiments are managed using W\&B \citep{wandb}.} on three main questions:
1) Can CEN identify controlled effects at pixel-level? i.e. can it produce an accurate segmentation mask?
2) Some applications may not require pixel-level precision; we asses CEN on predicting controlled effects at attribute-level from both pixel masks and latent representations.
3) Can RL agents benefit from using CEN as intrinsic motivator?

\begin{figure}[t]
    \centering
    \includegraphics[width=0.8\linewidth]{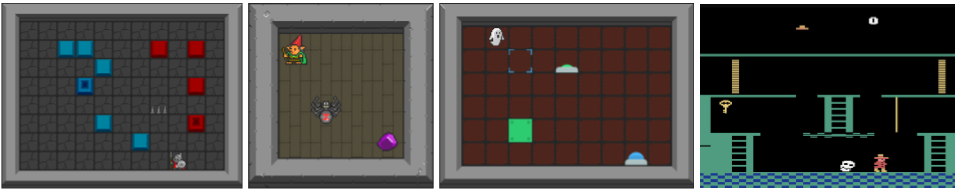}
    \caption{Suite of environments used for the experiments. From left/right top/bottom: Clusters, Spiders, Lights, and Montezuma's Revenge (MZR).}
    \label{fig:environments}
\end{figure}

\textbf{Environments:} we use multiple environments (Fig. \ref{fig:environments}) to answer the above questions, each showcasing a different aspect of what can be controlled.
These environments are based on Griddly \citep{bamford2020griddly} and Atari ALE \citep{bellemare2012ale}, both using the Gym interface \citep{brockman2016openai}.
More details of these environments are given in each experiment and appendix.

\textbf{Baselines:} We use Attentive Dynamics Model (ADM) \citep{Choi2019} in experiments 1) and 2).
ADM is an action-prediction model based on an spatial attention mechanism. It works by predicting the action performed on individual image patches. Then, a spatial attention mechanism selects a sparse set of patches to use when making a final prediction of the action. The masks produced by the attention mechanism are considered controlled aspects of the environment. Although ADM does not produce pixel level masks, only at patch level, to our knowledge ADM is the most competitive method to provide pixel-level information about controlled aspects.
For 2) and 3) we rely on Never Give Up (NGU) \citep{badia2020}, the current SOTA for exploration, and its inverse model.

\textbf{Implementation:} CEN is implemented as an encoder-decoder architecture with 2D convolutional layers and ReLU activation functions; the normal and controlled branches are implemented with linear layers. Additionally decoder weights are shared.
Throughout the experiments we use the same neural networks and hyperparameters unless specified otherwise.
Our implementation of ADM uses the same architecture and hyperparameters proposed in \citet{Choi2019}.
See appendix for more details on the architecture and hyperparameters.

\subsection{Controlled effects at pixel-level}
\label{sec:experiments_pixel}
This set of experiments explores CEN's ability to identify pixels corresponding to controlled entities.
Although CEN computes the magnitude and direction of the effects, we create a binary mask by setting a threshold for the predicted controlled effects (see exact details in appendix \ref{app:envs}).
We report pixel F1 scores between ground truth and predicted binary masks.
The network is trained to minimize Eq. \ref{eq:loss} using the ADAM optimizer \citep{KingmaB14} and 500K samples of the form $(o, a, e_p^a)$ collected using a random policy.

\subsubsection{Controlled vs uncontrolled effects}
Here we use the Spiders environment to evaluate CEN's ability to disentangle controlled from uncontrolled effects. This environment has two main entities, the agent and a spider.
The controlled masks must only focus on the agent and ignore the spider.

\begin{figure}[b]
  \begin{subfigure}{0.45\textwidth}
    \includegraphics[width=\textwidth]{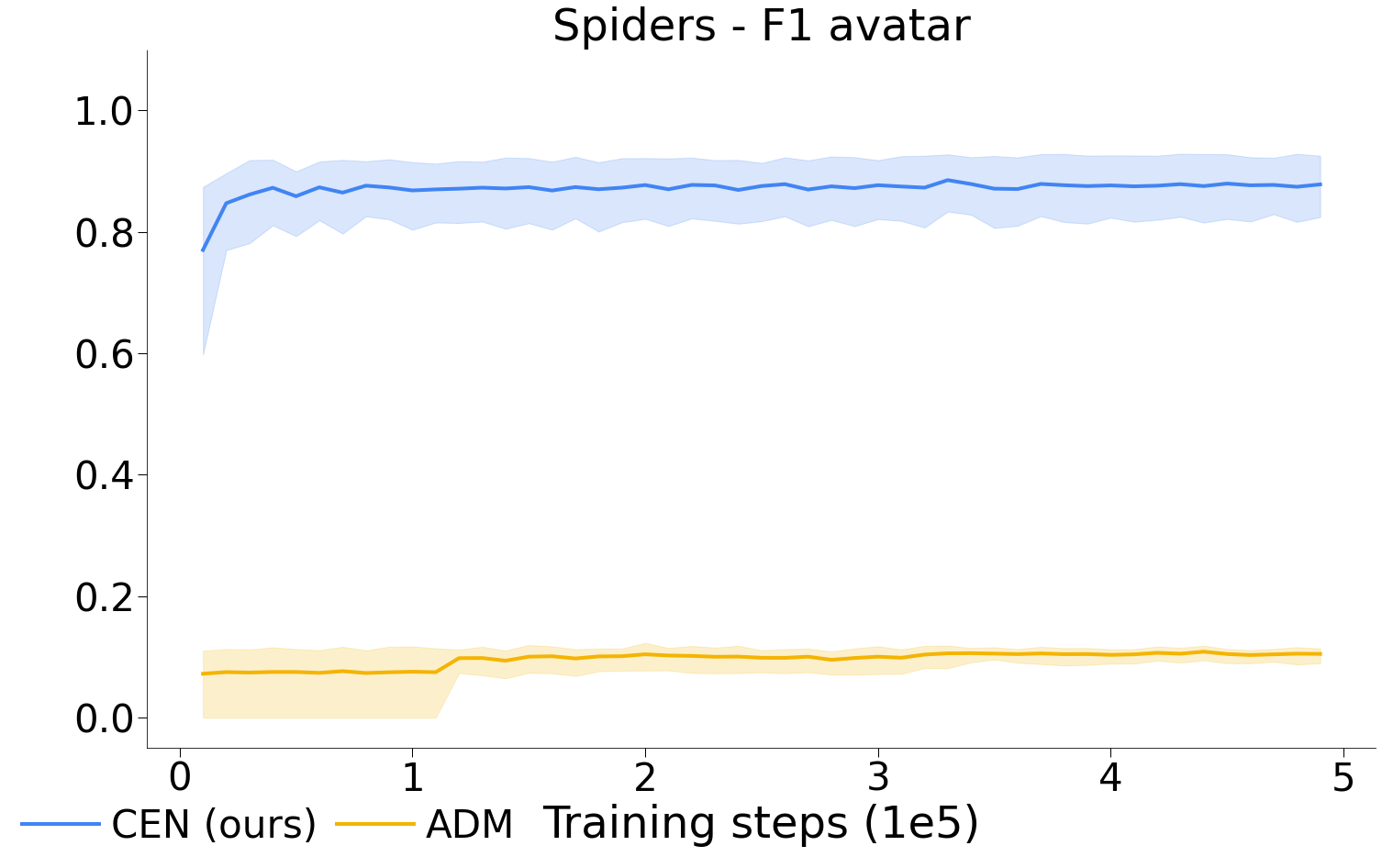}
    \caption{}
    \label{fig:results_disentanglement}
  \end{subfigure}
  \hfill
  \begin{subfigure}{0.45\textwidth}
    \includegraphics[width=\textwidth]{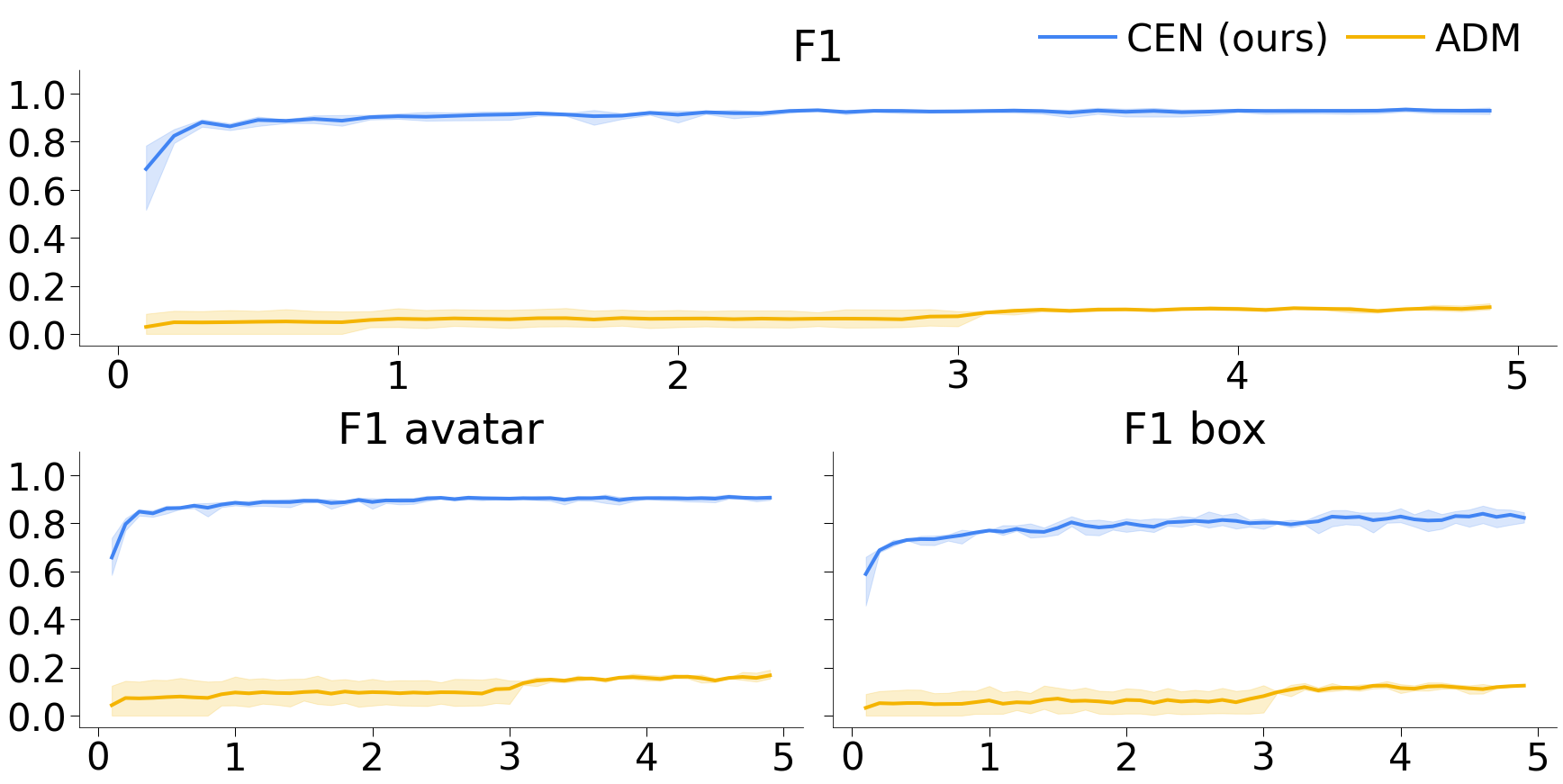}
    \caption{}
    \label{fig:results_close_effects}
  \end{subfigure}
  \caption{a) CEN can correctly disentangle the agent from the randomly moving objects. b) Clusters environment where CEN is able to model not just the agent but also the movement of boxes.}
\end{figure}

Fig. \ref{fig:results_disentanglement} shows the pixel F1 score for our model and the baseline. CEN is able to correctly disentangle controlled effects and can produce accurate masks.
Although our implementation of ADM can predict the agent's action with $88\%$ accuracy, it is not capable of modeling the agent's controlled pixels.
We conjecture that this is due to ADM's sparse softmax mechanism; nonetheless this behavior persisted when increasing its entropy weight which should produce more dense masks.

\subsubsection{Nearby controlled effects}
Models based on action prediction are expected to work well on aspects related to the agent.
For example, if an agent moves a box due to moving right; the box's movement is also controlled.
It is unclear why these models would pay attention to the box since just knowing where the agent is, suffices to predict the chosen action.
CEN's controlled branch, on the other hand, is motivated to model the box's effect since the normal branch would predict that the box stays where it is.
We call \say{nearby} controlled effect to an effect that happens adjacent the agent, like the box's movement.
To evaluate CEN on this kind of effects we use the Clusters environment where an agent needs to move colored boxes to their corresponding fixed colored blocks.
Fig. \ref{fig:results_close_effects} shows that CEN can precisely model effects on the agent and boxes. We breakdown individual effects to account for the class imbalance between the agent's movement and the boxes. CEN seems to make more mistakes with boxes than the agent but nonetheless, it can consistently model both.

\subsubsection{Faraway controlled effects}
In contrast to the previous experiment, we want to evaluate if CEN can model distant effects, i.e. effects that are reasonably far away from the agent's location.
In this case, we use the Lights environment. Here the environment presents two buttons of different color that, when pressed, turn on their corresponding light.
Lights are relatively far away from their corresponding buttons thus making it difficult to model them.
As show in Fig. \ref{fig:results_distant_effects}, CEN is able to model this kind of effects.

\subsubsection{Controlled effects in Montezuma's Revenge}
\begin{figure}[t]
  \begin{subfigure}{0.48\textwidth}
    \includegraphics[width=\textwidth]{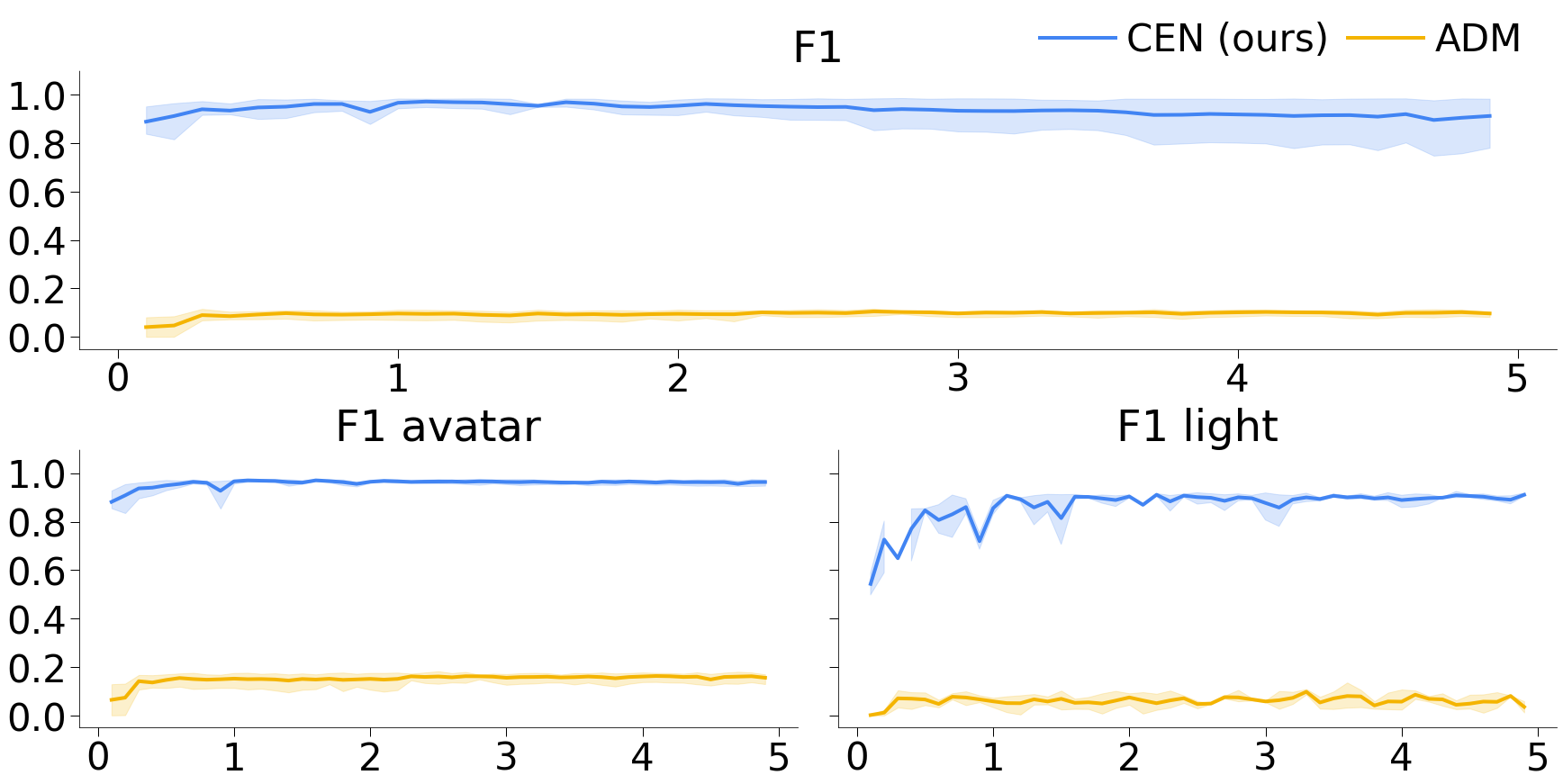}
    \caption{}
    \label{fig:results_distant_effects}
  \end{subfigure}
  \hfill
  \begin{subfigure}{0.48\textwidth}
    \includegraphics[width=\textwidth]{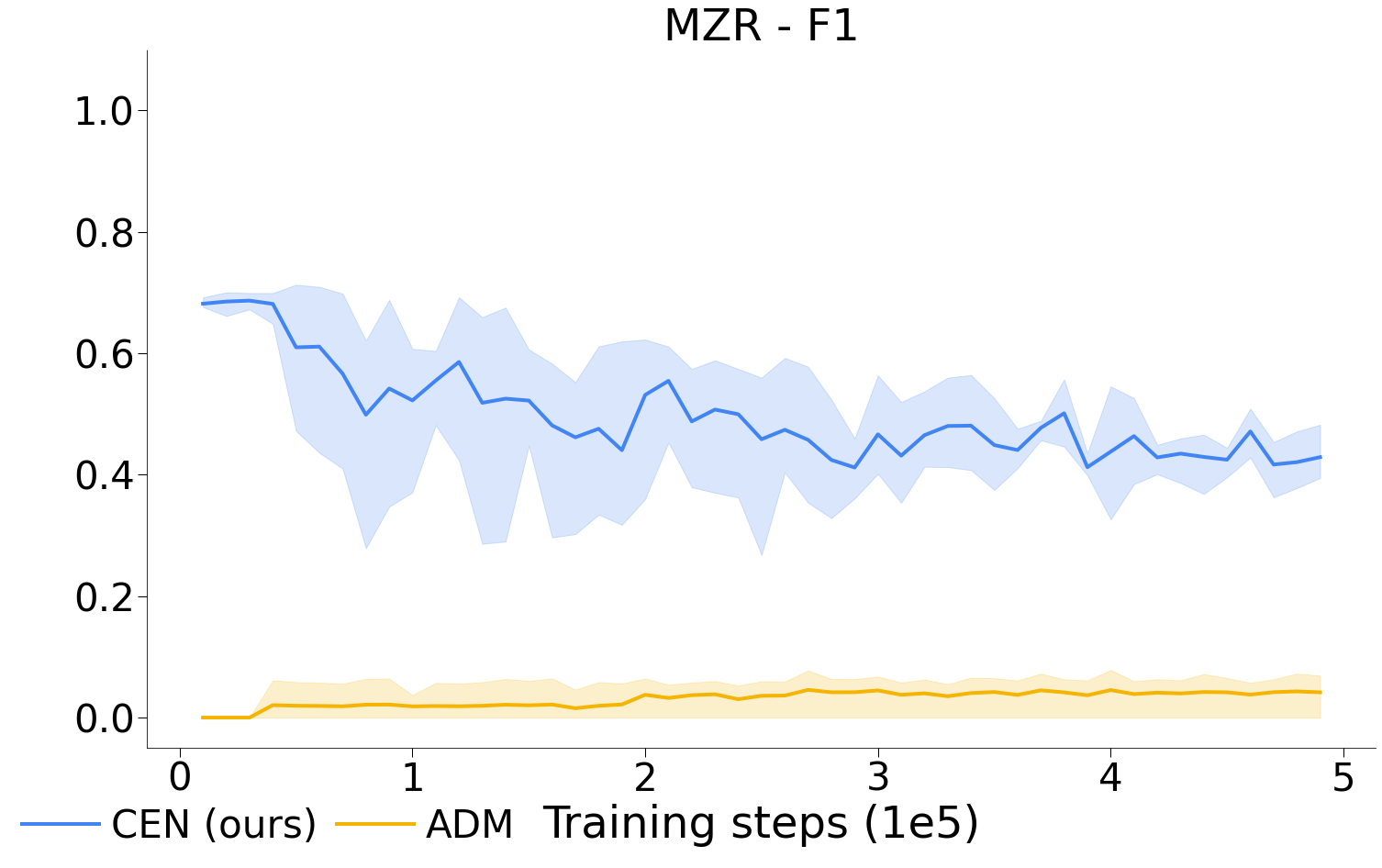}
    \caption{}
    \label{fig:results_mzr_effects}
  \end{subfigure}
  \caption{F1 score on the Lights (a) and MZR (b) environments. CEN outperforms the baseline even in an environment with more complex dynamics and features.}
\end{figure}

This last pixel-level experiment evaluates CEN on Montezuma's Revenge (MZR) environment.
Although agents in Atari environments have limited control over the environment, the relatively complex dynamics of Montezuma's Revenge makes it a challenging test-bed.
Results shown in Fig. \ref{fig:results_mzr_effects} indicate that CEN can also model controlled effects.

Unlike previous environments, CEN's F1 score behaves differently due to two main reasons:
1) \textit{F1 decreases over time:} the F1 score is extremely sensitive for empty ground truth, a single pixel marked as controlled takes the score from 1 to 0. At the beginning of training, CEN produces empty masks correctly but performs poorly at controlled effects, which makes the F1 score artificially high. Over time it improves at controlled effects but starts marking very few pixels as controlled for effects where nothing is controlled.
2) \textit{F1 is lower:} the sensitivity of the F1 score also affects CEN's performance on controlled effects since pixels outside the ground truth are penalized the same no matter how close to the ground truth they are.
Appendix \ref{app:masks} includes additional masks showing that CEN fails by pixels close to the controlled object instead of identifying the wrong object as controlled.

\begin{figure}[b]
  \centering
  \includegraphics[width=0.9\textwidth]{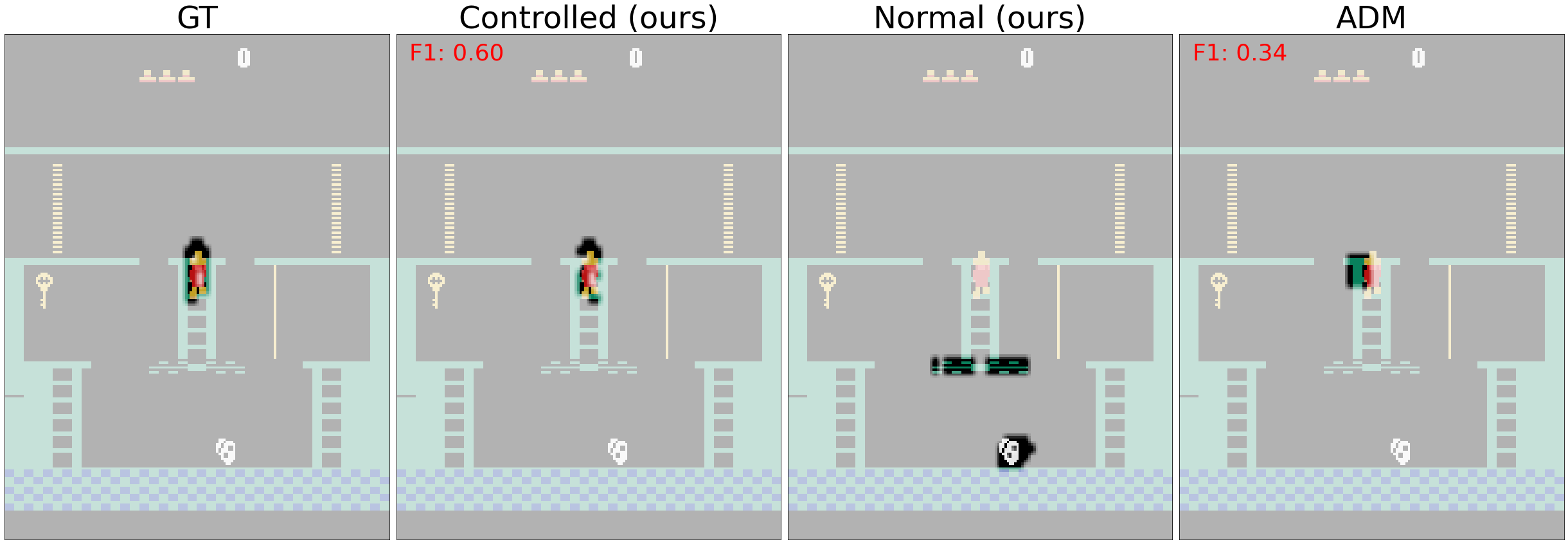}
  \caption{Example masks for Montezuma's Revenge. Additional masks are included in appendix \ref{app:masks}.}
  \label{fig:mask_main_mzr}
\end{figure}

\subsection{Controlled effects at attribute-level}
\label{sec:experiments_attribute}
In some cases requiring pixel-level precision can be excessive. The following experiments analyze how different representations can predict effects on attributes from the environment's state, e.g. changes on the agent's $(x,y)$ location or if a light turned on.
To this end, we use a probing technique similar to the one described in \citet{Anand2019}.
This approach trains a classifier per each attribute of interest in the environment's state using frozen versions of trained networks to produce the classifiers inputs.
More specifically we use two different sources for the probing classifiers, pixel-level masks (as in the previous section) or the model's latent representation.

For the first case, we produce a binary mask, as in the previous experiments, to occlude perceived effects and use these to train each probing classifier. Classifiers have to predict if there was a positive, negative or none effect. Note that the classifiers need to predict any effect, not just controlled.
Thus, the classifier should only be able to predict accurately controlled attributes such as the agent's position but should fail at predicting uncontrollable effects like the location of the spider or skull.
We use a random policy to collect a dataset of 35K samples of the form $(m * e_p, y)$ where $m$ is the mask produced by the model and $y$ is the ground truth class. Each dataset is split into a typical 70/20/10, allowing a $20\%$ class imbalance. We report F1 score of each attribute on the test set.

\subsubsection{Latent probing from pixels}
The results in Table \ref{tbl:results_attribute}(first block) indicate that CEN significantly outperforms the baseline when predicting controlled effects for state's attributes, and thus modeling controlled effects accurately. Furthermore, for both Spiders and Montezuma's Revenge environments the model cannot predict the uncontrolled effects, as expected.
Even though ADM's action prediction accuracy was high ($\sim88\%$) on every environment, it is not able to consistently predict controlled effects at attribute-level.

\subsubsection{Latent probing from latent representations}
In this case, we train classifiers using a latent representation instead of pixels. We use the latent representation from CEN's controlled branch ($h_c$). It is unclear how to create a latent representation from ADM, so we use an inverse model. The features of current and next observations are concatenated to create a latent representation of what is controlled.
As before, we train a linear probe to predict controlled attributes from the latent space of these models. The probe predicts changes on the x, y and direction (if applies) for agent, spider, skull and boxes; and on/off for lights and buttons.

\begin{table}
\begin{center}
\begin{small}
\begin{sc}
\begin{tabular}{llcccc}
\toprule
& & \multicolumn{2}{c}{F1 Pixels} & \multicolumn{2}{c}{F1 Latent} \\
\cmidrule(lr){3-6}
Environment & Attribute & CEN (ours) & ADM & CEN (ours) & Inverse\\
\midrule
\multirow{2}{*}{Spiders} & Agent & \textbf{1.0}$\pm$\textbf{0.00} & 0.47$\pm$0.23 & \textbf{0.97}$\pm$\textbf{0.01} & 0.67$\pm$0.05\\
                         & Spider $\downarrow$ & 0.35$\pm$0.03 & \textbf{0.25}$\pm$\textbf{0.03} & \textbf{0.41}$\pm$\textbf{0.01} & 0.44$\pm$0.02\\
\addlinespace
\hline
\addlinespace
\multirow{2}{*}{Clusters} & Agent & \textbf{0.76}$\pm$\textbf{0.41} & 0.28$\pm$0.08 & \textbf{0.97}$\pm$\textbf{0.01} & 0.56$\pm$0.09\\
                          & Box & \textbf{0.78}$\pm$\textbf{0.37} & 0.32$\pm$0.19 & \textbf{0.95}$\pm$\textbf{0.02} & 0.77$\pm$0.00\\
\addlinespace
\hline
\addlinespace
\multirow{3}{*}{Lights} & Agent & \textbf{0.97}$\pm$\textbf{0.01} & 0.33$\pm$0.15 & \textbf{1.0}$\pm$\textbf{0.01} & 0.84$\pm$0.08\\
                        & Button & \textbf{0.93}$\pm$\textbf{0.05} & 0.33$\pm$0.01 & \textbf{0.99}$\pm$\textbf{0.0} & \textbf{0.99}$\pm$\textbf{0.0}\\
                        & Light & \textbf{0.93}$\pm$\textbf{0.04} & 0.41$\pm$0.14 & \textbf{1.0}$\pm$\textbf{0.0} & 0.99$\pm$0.0\\
\addlinespace
\hline
\addlinespace
\multirow{2}{*}{MZR} & Agent & \textbf{0.66}$\pm$\textbf{0.08} & 0.42$\pm$0.23 & \textbf{0.91}$\pm$\textbf{0.02} & 0.88$\pm$0.02\\
                     & Skull $\downarrow$ & \textbf{0.19}$\pm$\textbf{0.03} & 0.20$\pm$0.08 & \textbf{0.61}$\pm$\textbf{0.03} & 0.61$\pm$0.04\\
\bottomrule
\end{tabular}
\end{sc}
\end{small}
\end{center}
\caption{F1 score for the state attributes when predicted from pixel or latent space. Note that the lower the score for Skull and Spider the better.}
\label{tbl:results_attribute}
\end{table}

Table \ref{tbl:results_attribute} (second block) show that CEN improves on the baseline's performance.
Although the inverse model is closer to CEN's performance than ADM, it still has difficulties predicting the agent and box changes in location.
The reason of having high score in Skull is that the Skull is still only when the agent dies, making it easy to predict from the agent's position. Removing this event leads to a score of $\sim0.35$ for both models.

\subsection{CEN as intrinsic motivator}
\label{sec:experiments_rl}
We have shown that CEN can learn controlled effects in an unsupervised manner. Here we showcase the use of this ability as an intrinsic motivator of a reinforcement learning agent.
We consider two tasks, an empty environment without any extrinsic reward where the agent can only control itself and Clusters.
The RL agent is implemented using PPO (vanilla). Additionally, PPO is augmented with the exploration bonus proposed in Never Give Up (NGU + Inverse) \citep{badia2020}. NGU is composed of two modules for computing episodic and life-long rewards. For simplicity the following experiments only use the episodic module which in NGU consists of a count-based method using an episodic memory to approximate the number of times an agent visited each state, and an inverse model to identify controlled states. We replace the inverse model with CEN (NGU + CEN) and use the latent representation of the controlled branch to compute NGU's episodic reward. In this experiments, CEN is trained along the PPO policy using experiences collected with the same policy.

\textbf{Empty environment:} the goal of this environment is to showcase how each NGU variant rewards controlled events. In this environment, only the agent's movement is controlled; thus, rewarding for controlling the agent's location should promote a uniform exploration of the environment since once a location is visited, it should not be rewarded as much as visiting a new location. We hypothesize that an inverse model will create similar representations for the same action disregarding the location where it was taken. This should impair exploration since the reward will be similar regardless of where the action is performed.

\begin{figure}
  \begin{subfigure}{0.4\textwidth}
    \includegraphics[width=\textwidth]{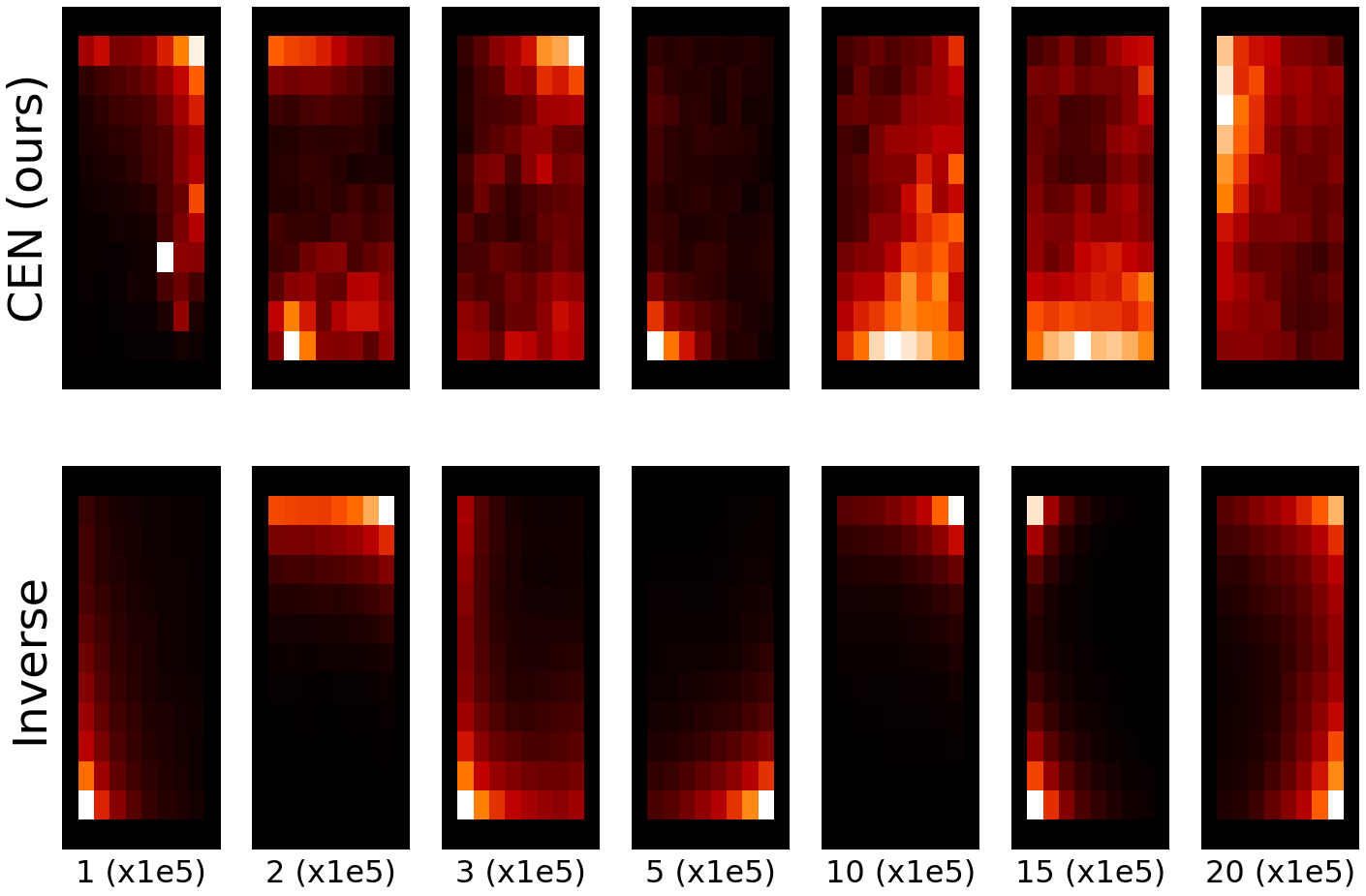}
    \caption{}
    \label{fig:results_rl_empty}
  \end{subfigure}
  \hfill
  \begin{subfigure}{0.58\textwidth}
    \includegraphics[width=\textwidth]{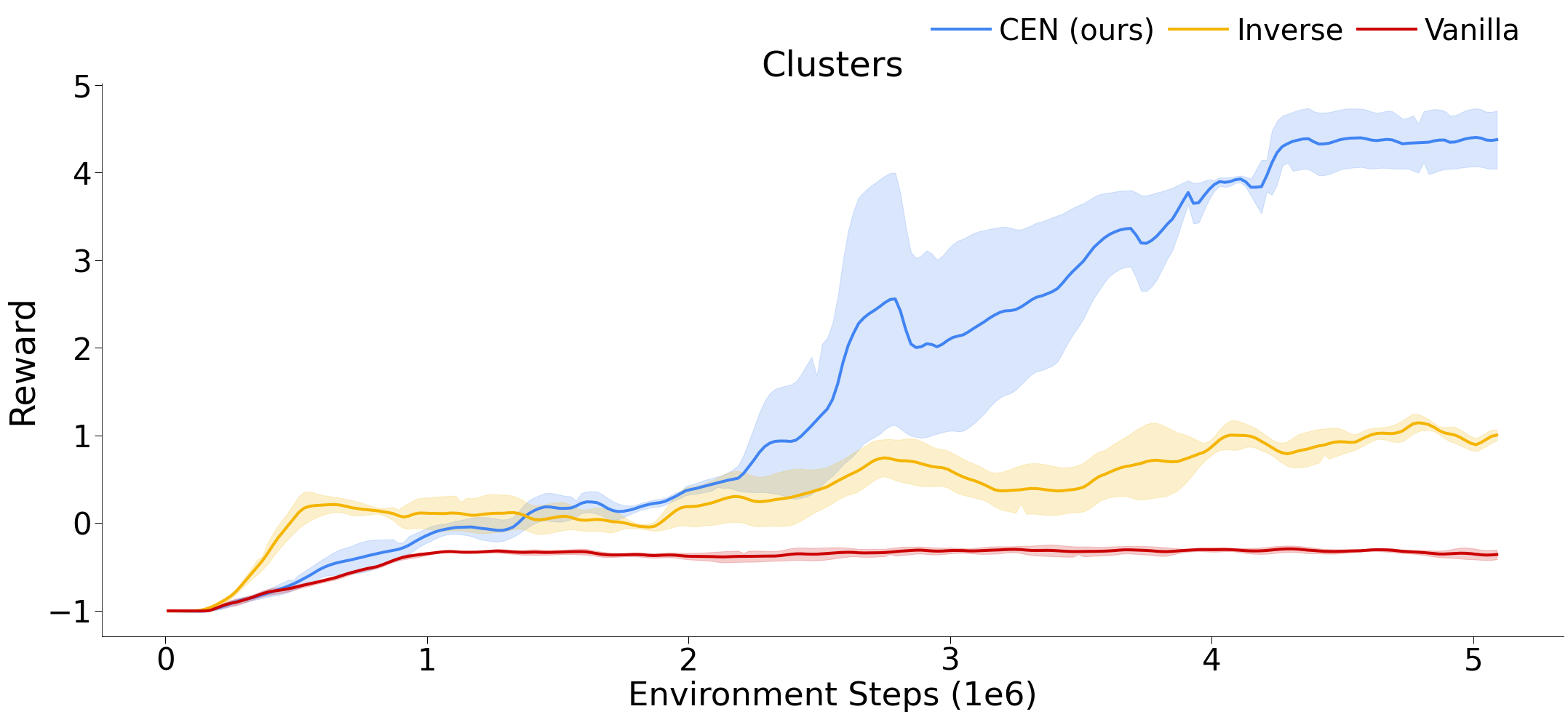}
    \caption{}
    \label{fig:results_rl_clusters}
  \end{subfigure}
  \caption{a) State visitation maps at different points of training of the Empty environment. CEN values different locations similarly, and consequently, the agent learns to explore states more uniformly. The inverse model encourages going to walls where predicting the action is hard. b) CEN promotes the movement of boxes and consequently faster learning}
\end{figure}

Fig. \ref{fig:results_rl_empty} shows that agent trained with the inverse model hogs walls and corners. This is because is hard to predict the action near unmovable objects, leading to higher value. Contrarily, CEN promotes different locations more uniformly and the agent learns to explore better the environment.

\textbf{Clusters environment:} A more challenging environment is Clusters, where the agent needs to move colored boxes to their respective colored blocks. This environment provides a reward at the end of the episode corresponding to the total number of boxes correctly placed. Results are provided in Fig. \ref{fig:results_rl_clusters}. Due to the sparsity of the reward PPO does not learn a correct behavior in the given time. Similarly, NGU + Inverse learns to place one box but fails to learn a general behavior to solve the task. Conversely, NGU + CEN quickly learns to move boxes leading to a high extrinsic reward. 

\section{Related work}
\label{sec:related_work}

\textbf{Intrinsic motivators:}
A popular way of introducing behavioral biases in RL agents is the use of intrinsic motivators \citep{Singh2005, mohamed2015variational}.
These motivators can promote different types of exploration, from observational surprise \citep{Burda2018} to control seeking agents \citep{pathak2017, Choi2019}.
Methods in the latter category have shown extremely good results achieving SOTA in important benchmarks.
\citet{Choi2019} proposed Attentive Dynamics Model (ADM), an attention based method that discovers controlled elements in the environment and rewards the agent for discovering them. This method and its extension \citep{song2019megareward} showed SOTA in Montezuma's Revenge.
\citet{badia2020} combined control and observational surprise to promote exploration. Their method uses an episodic memory with an inverse model to promote the discovery of controlled effects and Random Network Distillation \citep{Burda2018} to promote long term progress; again achieving SOTA in Atari's hard exploration environments.
These methods show the importance of identifying what an agent can control.

\textbf{Causality in deep reinforcement learning:}
Causality is central to humans; we think in terms of cause-effect.
A similar method to Blame was proposed in \citet{chattopadhyay2019}, where they use causal attribution methods to analyzed the effect of inputs on a neural network's outputs.
Recent work has introduced causality into deep reinforcement learning \citep{Foerster_2018, Buesing2018,Jaques2018,Dasgupta2019,Goyal2019,FeiFei2019, Madumal2020} showing that this is a promising avenue for the training of agents.
\citet{corcoll2020disentangling} proposed an attribution method to learn temporal abstractions for object-centric hierarchical RL.
\citet{Bellemare2012InvestigatingCA} compute controllable aspects of the environment by generating a mask with all possible controllable areas of an image and uses it as part of the policy's input.
In this work, we identify the controlled effects of individual actions using causal concepts of normality and blame.

\section{Conclusions and limitations}
\label{sec:conclusions}
This work proposes a fully unsupervised approach to this problem named Controlled Effect Network (CEN).
CEN creates a normative world using counterfactuals and compares what actually happened with what normally would happen to attribute changes on the environment to the agent.
The presented experiments show that, despite being unsupervised, this method precisely identifies controlled effects.
Furthermore, CEN is showcased as intrinsic motivator for RL agents where results suggest that a more targeted exploration leads to substantially better policies.

\textbf{Limitations}

\textit{Normality:} although we propose a measure of normality in Eq. 5, this is far from ideal. We believe the way humans see normality is context dependent and should be learned instead of a fixed function. This is an active research area by psychologist looking to underpin human causal judgment.


\textit{State instead of observations:} our current formulation of ICE uses perceived effects as opposed to states. When CEN indicates that some change is controlled this is not (necessarily) equivalent to stating that the objects represented by those pixels are controlled. An exciting avenue to explore is combining CEN with a state representation model (e.g. LSTM or RSSM by \citet{hafner2019learning}) where changes would happen at state level.

\textit{Reliance on policy:} since Eq. \ref{eq:controlled_effects} does not depend on the policy, CEN needs diverse data for each action to approximate it, ideally from a random policy. This important problem is not specific to CEN, any forward model needs a policy that explores multiple actions to provide accurate predictions.


\section*{Acknowledgments}
The authors would like to thank Jaan Aru and Meelis Kull for insightful comments on the manuscript. This work was supported by the University of Tartu ASTRA Project PER ASPERA, financed by the European Regional Development Fund and the University of Tartu HPC.

\bibliography{egbib}

\begin{thebibliography}{49}
\providecommand{\natexlab}[1]{#1}
\providecommand{\url}[1]{\texttt{#1}}
\expandafter\ifx\csname urlstyle\endcsname\relax
  \providecommand{\doi}[1]{doi: #1}\else
  \providecommand{\doi}{doi: \begingroup \urlstyle{rm}\Url}\fi

\bibitem[Anand et~al.(2019)Anand, Racah, Ozair, Bengio, C{\^{o}}t{\'{e}}, and
  Hjelm]{Anand2019}
Ankesh Anand, Evan Racah, Sherjil Ozair, Yoshua Bengio, Marc-Alexandre
  C{\^{o}}t{\'{e}}, and R~Devon Hjelm.
\newblock {Unsupervised State Representation Learning in Atari}.
\newblock 2019.

\bibitem[Badia et~al.(2020{\natexlab{a}})Badia, Piot, Kapturowski, Sprechmann,
  Vitvitskyi, Guo, and Blundell]{badia2020agent57}
Adrià~Puigdomènech Badia, Bilal Piot, Steven Kapturowski, Pablo Sprechmann,
  Alex Vitvitskyi, Daniel Guo, and Charles Blundell.
\newblock Agent57: Outperforming the atari human benchmark, 2020{\natexlab{a}}.

\bibitem[Badia et~al.(2020{\natexlab{b}})Badia, Sprechmann, Vitvitskyi, Guo,
  Piot, Kapturowski, Tieleman, Arjovsky, Pritzel, Bolt, and
  Blundell]{badia2020}
Adrià~Puigdomènech Badia, Pablo Sprechmann, Alex Vitvitskyi, Daniel Guo,
  Bilal Piot, Steven Kapturowski, Olivier Tieleman, Martín Arjovsky, Alexander
  Pritzel, Andew Bolt, and Charles Blundell.
\newblock Never give up: Learning directed exploration strategies,
  2020{\natexlab{b}}.

\bibitem[Baker et~al.(2019)Baker, Kanitscheider, Markov, Wu, Powell, McGrew,
  and Mordatch]{baker2019}
Bowen Baker, Ingmar Kanitscheider, Todor Markov, Yi~Wu, Glenn Powell, Bob
  McGrew, and Igor Mordatch.
\newblock Emergent tool use from multi-agent autocurricula, 2019.

\bibitem[Bamford et~al.(2020)Bamford, Huang, and Lucas]{bamford2020griddly}
Chris Bamford, Shengyi Huang, and Simon Lucas.
\newblock Griddly: A platform for ai research in games, 2020.

\bibitem[Bellemare et~al.(2012{\natexlab{a}})Bellemare, Naddaf, Veness, and
  Bowling]{bellemare2012ale}
Marc~G. Bellemare, Yavar Naddaf, Joel Veness, and Michael Bowling.
\newblock The arcade learning environment: An evaluation platform for general
  agents.
\newblock \emph{Journal of Artificial Intelligence Research}, Vol. 47:\penalty0
  253--279, 2012{\natexlab{a}}.
\newblock \doi{10.1613/jair.3912}.
\newblock URL \url{http://arxiv.org/abs/1207.4708}.
\newblock cite arxiv:1207.4708.

\bibitem[Bellemare et~al.(2012{\natexlab{b}})Bellemare, Veness, and
  Bowling]{Bellemare2012InvestigatingCA}
Marc~G. Bellemare, J.~Veness, and Michael Bowling.
\newblock Investigating contingency awareness using atari 2600 games.
\newblock In \emph{AAAI}, 2012{\natexlab{b}}.

\bibitem[Biewald(2020)]{wandb}
Lukas Biewald.
\newblock Experiment tracking with weights and biases, 2020.
\newblock URL \url{https://www.wandb.com/}.
\newblock Software available from wandb.com.

\bibitem[Brockman et~al.(2016)Brockman, Cheung, Pettersson, Schneider,
  Schulman, Tang, and Zaremba]{brockman2016openai}
Greg Brockman, Vicki Cheung, Ludwig Pettersson, Jonas Schneider, John Schulman,
  Jie Tang, and Wojciech Zaremba.
\newblock Openai gym, 2016.
\newblock URL \url{http://arxiv.org/abs/1606.01540}.
\newblock cite arxiv:1606.01540.

\bibitem[Buesing et~al.(2018)Buesing, Weber, Zwols, Racaniere, Guez, Lespiau,
  and Heess]{Buesing2018}
Lars Buesing, Theophane Weber, Yori Zwols, Sebastien Racaniere, Arthur Guez,
  Jean-Baptiste Lespiau, and Nicolas Heess.
\newblock Woulda, coulda, shoulda: Counterfactually-guided policy search, 2018.

\bibitem[Burda et~al.(2018)Burda, Edwards, Storkey, and Openai]{Burda2018}
Yuri Burda, Harrison Edwards, Amos Storkey, and Oleg~Klimov Openai.
\newblock {Exploration by Random Network Distillation}.
\newblock 2018.

\bibitem[Chattopadhyay et~al.(2019)Chattopadhyay, Manupriya, Sarkar, and
  Balasubramanian]{chattopadhyay2019}
Aditya Chattopadhyay, Piyushi Manupriya, Anirban Sarkar, and Vineeth~N
  Balasubramanian.
\newblock Neural network attributions: A causal perspective.
\newblock In Kamalika Chaudhuri and Ruslan Salakhutdinov, editors,
  \emph{Proceedings of the 36th International Conference on Machine Learning},
  volume~97 of \emph{Proceedings of Machine Learning Research}, pages 981--990.
  PMLR, 09--15 Jun 2019.
\newblock URL \url{http://proceedings.mlr.press/v97/chattopadhyay19a.html}.

\bibitem[Chentanez et~al.(2005)Chentanez, Barto, and Singh]{ChentanezIntrinsic}
Nuttapong Chentanez, Andrew Barto, and Satinder Singh.
\newblock Intrinsically motivated reinforcement learning.
\newblock In L.~Saul, Y.~Weiss, and L.~Bottou, editors, \emph{Advances in
  Neural Information Processing Systems}, volume~17, pages 1281--1288. MIT
  Press, 2005.
\newblock URL
  \url{https://proceedings.neurips.cc/paper/2004/file/4be5a36cbaca8ab9d2066debfe4e65c1-Paper.pdf}.

\bibitem[Choi et~al.(2019)Choi, Guo, Moczulski, Oh, Wu, Norouzi, and
  Lee]{Choi2019}
Jongwook Choi, Yijie Guo, Marcin Moczulski, Junhyuk Oh, Neal Wu, Mohammad
  Norouzi, and Honglak Lee.
\newblock {Contingency-Aware Exploration in Reinforcement Learning}.
\newblock \emph{ICLR}, pages 1--20, 2019.

\bibitem[Corcoll and Vicente(2020)]{corcoll2020disentangling}
Oriol Corcoll and Raul Vicente.
\newblock Disentangling causal effects for hierarchical reinforcement learning,
  2020.

\bibitem[Cushman et~al.(2008)Cushman, Knobe, and
  Sinnott-Armstrong]{Cushman2008}
Fiery Cushman, Joshua Knobe, and Walter Sinnott-Armstrong.
\newblock Moral appraisals affect doing/allowing judgments.
\newblock \emph{Cognition}, 108\penalty0 (1):\penalty0 281--289, July 2008.
\newblock \doi{10.1016/j.cognition.2008.02.005}.
\newblock URL \url{https://doi.org/10.1016/j.cognition.2008.02.005}.

\bibitem[Dasgupta et~al.(2019)Dasgupta, Wang, Chiappa, Mitrovic, Ortega,
  Raposo, Hughes, Battaglia, Botvinick, and Kurth-Nelson]{Dasgupta2019}
Ishita Dasgupta, Jane Wang, Silvia Chiappa, Jovana Mitrovic, Pedro Ortega,
  David Raposo, Edward Hughes, Peter Battaglia, Matthew Botvinick, and Zeb
  Kurth-Nelson.
\newblock Causal reasoning from meta-reinforcement learning, 2019.

\bibitem[Espeholt et~al.(2018)Espeholt, Soyer, Munos, Simonyan, Mnih, Ward,
  Doron, Firoiu, Harley, Dunning, Legg, and Kavukcuoglu]{espeholt2018impala}
Lasse Espeholt, Hubert Soyer, Remi Munos, Karen Simonyan, Volodymir Mnih, Tom
  Ward, Yotam Doron, Vlad Firoiu, Tim Harley, Iain Dunning, Shane Legg, and
  Koray Kavukcuoglu.
\newblock Impala: Scalable distributed deep-rl with importance weighted
  actor-learner architectures, 2018.

\bibitem[Foerster et~al.(2018)Foerster, Farquhar, Afouras, Nardelli, and
  Whiteson]{Foerster_2018}
Jakob Foerster, Gregory Farquhar, Triantafyllos Afouras, Nantas Nardelli, and
  Shimon Whiteson.
\newblock Counterfactual multi-agent policy gradients.
\newblock \emph{Proceedings of the AAAI Conference on Artificial Intelligence},
  32\penalty0 (1), Apr. 2018.
\newblock URL \url{https://ojs.aaai.org/index.php/AAAI/article/view/11794}.

\bibitem[Fujita et~al.(2021)Fujita, Nagarajan, Kataoka, and
  Ishikawa]{Fujita2021}
Yasuhiro Fujita, Prabhat Nagarajan, Toshiki Kataoka, and Takahiro Ishikawa.
\newblock Chainerrl: A deep reinforcement learning library.
\newblock \emph{Journal of Machine Learning Research}, 22\penalty0
  (77):\penalty0 1--14, 2021.
\newblock URL \url{http://jmlr.org/papers/v22/20-376.html}.

\bibitem[Gerstenberg and Lagnado(2014)]{gerstenberg2014}
T.~Gerstenberg and D.~A. Lagnado.
\newblock Attributing responsibility: Actual and counterfactual worlds.
\newblock In Joshua Knobe, Tania Lombrozo, and Shaun Nichols, editors,
  \emph{Oxford Studies in Experimental Philosophy}, volume~1, pages 91--130.
  Oxford University Press, 2014.

\bibitem[Goyal et~al.(2019)Goyal, Sodhani, Binas, Peng, Levine, and
  Bengio]{Goyal2019}
Anirudh Goyal, Shagun Sodhani, Jonathan Binas, Xue~Bin Peng, Sergey Levine, and
  Yoshua Bengio.
\newblock {Reinforcement Learning with Competitive Ensembles of
  Information-Constrained Primitives}.
\newblock pages 1--21, 2019.

\bibitem[Grinfeld et~al.(2020)Grinfeld, Lagnado, Gerstenberg, Woodward, and
  Usher]{Grinfeld2020}
Guy Grinfeld, David Lagnado, Tobias Gerstenberg, James~F. Woodward, and Marius
  Usher.
\newblock Causal responsibility and robust causation.
\newblock \emph{Frontiers in Psychology}, 11:\penalty0 1069, 2020.
\newblock ISSN 1664-1078.
\newblock \doi{10.3389/fpsyg.2020.01069}.
\newblock URL
  \url{https://www.frontiersin.org/article/10.3389/fpsyg.2020.01069}.

\bibitem[Gulcehre et~al.(2020)Gulcehre, Paine, Shahriari, Denil, Hoffman,
  Soyer, Tanburn, Kapturowski, Rabinowitz, Williams, Barth-Maron, Wang,
  de~Freitas, and Team]{Gulcehre2020}
Caglar Gulcehre, Tom~Le Paine, Bobak Shahriari, Misha Denil, Matt Hoffman,
  Hubert Soyer, Richard Tanburn, Steven Kapturowski, Neil Rabinowitz, Duncan
  Williams, Gabriel Barth-Maron, Ziyu Wang, Nando de~Freitas, and Worlds Team.
\newblock Making efficient use of demonstrations to solve hard exploration
  problems.
\newblock In \emph{International Conference on Learning Representations}, 2020.
\newblock URL \url{https://openreview.net/forum?id=SygKyeHKDH}.

\bibitem[Hafner et~al.(2019)Hafner, Lillicrap, Fischer, Villegas, Ha, Lee, and
  Davidson]{hafner2019learning}
Danijar Hafner, Timothy Lillicrap, Ian Fischer, Ruben Villegas, David Ha,
  Honglak Lee, and James Davidson.
\newblock Learning latent dynamics for planning from pixels, 2019.

\bibitem[Halpern(2016)]{Halpern2016}
Joseph~Y. Halpern.
\newblock \emph{Actual Causality}.
\newblock The MIT Press, 2016.
\newblock ISBN 0262035022.

\bibitem[Halpern and Hitchcock(2014)]{Halpern2014}
Joseph~Y. Halpern and Christopher Hitchcock.
\newblock {Graded Causation and Defaults}.
\newblock \emph{The British Journal for the Philosophy of Science}, 66\penalty0
  (2):\penalty0 413--457, 04 2014.
\newblock ISSN 0007-0882.
\newblock \doi{10.1093/bjps/axt050}.

\bibitem[Harris et~al.(2020)Harris, Millman, van~der Walt, Gommers, Virtanen,
  Cournapeau, Wieser, Taylor, Berg, Smith, Kern, Picus, Hoyer, van Kerkwijk,
  Brett, Haldane, del R{'{\i}}o, Wiebe, Peterson, G{'{e}}rard-Marchant,
  Sheppard, Reddy, Weckesser, Abbasi, Gohlke, and Oliphant]{harris2020array}
Charles~R. Harris, K.~Jarrod Millman, St{'{e}}fan~J. van~der Walt, Ralf
  Gommers, Pauli Virtanen, David Cournapeau, Eric Wieser, Julian Taylor,
  Sebastian Berg, Nathaniel~J. Smith, Robert Kern, Matti Picus, Stephan Hoyer,
  Marten~H. van Kerkwijk, Matthew Brett, Allan Haldane, Jaime~Fern{'{a}}ndez
  del R{'{\i}}o, Mark Wiebe, Pearu Peterson, Pierre G{'{e}}rard-Marchant, Kevin
  Sheppard, Tyler Reddy, Warren Weckesser, Hameer Abbasi, Christoph Gohlke, and
  Travis~E. Oliphant.
\newblock Array programming with {NumPy}.
\newblock \emph{Nature}, 585\penalty0 (7825):\penalty0 357--362, September
  2020.
\newblock \doi{10.1038/s41586-020-2649-2}.
\newblock URL \url{https://doi.org/10.1038/s41586-020-2649-2}.

\bibitem[Hitchcock and Knobe(2009)]{Hitchcock2009}
Christopher Hitchcock and Joshua Knobe.
\newblock Cause and norm.
\newblock \emph{Journal of Philosophy}, 106\penalty0 (11):\penalty0 587--612,
  2009.
\newblock \doi{10.5840/jphil20091061128}.

\bibitem[Jaques et~al.(2018)Jaques, Lazaridou, Hughes, Gulcehre, Ortega,
  Strouse, Leibo, and de~Freitas]{Jaques2018}
Natasha Jaques, Angeliki Lazaridou, Edward Hughes, Caglar Gulcehre, Pedro~A.
  Ortega, DJ~Strouse, Joel~Z. Leibo, and Nando de~Freitas.
\newblock Social influence as intrinsic motivation for multi-agent deep
  reinforcement learning, 2018.

\bibitem[Kahneman and Miller(1986)]{kahneman_1986}
D.~Kahneman and D.~T. Miller.
\newblock Norm theory - comparing reality to its alternatives.
\newblock \emph{Psychol Rev Psychol Rev}, 93\penalty0 (2):\penalty0 136--153,
  1986.

\bibitem[Kapturowski et~al.(2019)Kapturowski, Ostrovski, Dabney, Quan, and
  Munos]{kapturowski2018}
Steven Kapturowski, Georg Ostrovski, Will Dabney, John Quan, and Remi Munos.
\newblock Recurrent experience replay in distributed reinforcement learning.
\newblock In \emph{International Conference on Learning Representations}, 2019.
\newblock URL \url{https://openreview.net/forum?id=r1lyTjAqYX}.

\bibitem[Kingma and Ba(2015)]{KingmaB14}
Diederik~P. Kingma and Jimmy Ba.
\newblock Adam: {A} method for stochastic optimization.
\newblock In Yoshua Bengio and Yann LeCun, editors, \emph{3rd International
  Conference on Learning Representations, {ICLR} 2015, San Diego, CA, USA, May
  7-9, 2015, Conference Track Proceedings}, 2015.
\newblock URL \url{http://arxiv.org/abs/1412.6980}.

\bibitem[Knobe and Fraser(2008)]{Knobe2008}
Joshua Knobe and Benjamin Fraser.
\newblock \emph{Causal judgment and moral judgment: Two experiments}, pages 441
  -- 447.
\newblock The MIT Press, United States of America, 2008.
\newblock ISBN 9780262195690.

\bibitem[Langenhoff et~al.(2019)Langenhoff, Wiegmann, Halpern, Tenenbaum, and
  Gerstenberg]{langenhoff2019}
Antonia~F Langenhoff, Alex Wiegmann, Joseph~Y Halpern, Joshua Tenenbaum, and
  Tobias Gerstenberg.
\newblock Predicting responsibility judgments from dispositional inferences and
  causal attributions, Sep 2019.
\newblock URL \url{psyarxiv.com/63zvw}.

\bibitem[Madumal et~al.(2020)Madumal, Miller, Sonenberg, and
  Vetere]{Madumal2020}
Prashan Madumal, Tim Miller, Liz Sonenberg, and Frank Vetere.
\newblock Explainable reinforcement learning through a causal lens.
\newblock \emph{Proceedings of the AAAI Conference on Artificial Intelligence},
  34\penalty0 (03):\penalty0 2493--2500, Apr. 2020.
\newblock \doi{10.1609/aaai.v34i03.5631}.
\newblock URL \url{https://ojs.aaai.org/index.php/AAAI/article/view/5631}.

\bibitem[Mohamed and Rezende(2015)]{mohamed2015variational}
Shakir Mohamed and Danilo~Jimenez Rezende.
\newblock Variational information maximisation for intrinsically motivated
  reinforcement learning, 2015.

\bibitem[Morris et~al.(2018)Morris, Phillips, Icard, Knobe, Gerstenberg, and
  Cushman]{morris2018}
Adam Morris, Jonathan~S Phillips, Thomas Icard, Joshua Knobe, Tobias
  Gerstenberg, and Fiery~A Cushman.
\newblock Causal judgments approximate the effectiveness of future
  interventions, Apr 2018.
\newblock URL \url{psyarxiv.com/nq53z}.

\bibitem[Nair et~al.(2019)Nair, Zhu, Savarese, and Fei-Fei]{FeiFei2019}
Suraj Nair, Yuke Zhu, Silvio Savarese, and Li~Fei-Fei.
\newblock Causal induction from visual observations for goal directed tasks,
  2019.

\bibitem[OpenAI et~al.(2019)OpenAI, Berner, Brockman, Chan, Cheung, Debiak,
  Dennison, Farhi, Fischer, Hashme, Hesse, Józefowicz, Gray, Olsson, Pachocki,
  Petrov, de~Oliveira~Pinto, Raiman, Salimans, Schlatter, Schneider, Sidor,
  Sutskever, Tang, Wolski, and Zhang]{openai2019dota}
OpenAI, Christopher Berner, Greg Brockman, Brooke Chan, Vicki Cheung,
  Przemys{\l}aw Debiak, Christy Dennison, David Farhi, Quirin Fischer, Shariq
  Hashme, Chris Hesse, Rafal Józefowicz, Scott Gray, Catherine Olsson, Jakub
  Pachocki, Michael Petrov, Henrique~Pondé de~Oliveira~Pinto, Jonathan Raiman,
  Tim Salimans, Jeremy Schlatter, Jonas Schneider, Szymon Sidor, Ilya
  Sutskever, Jie Tang, Filip Wolski, and Susan Zhang.
\newblock Dota 2 with large scale deep reinforcement learning, 2019.

\bibitem[Paszke et~al.(2019)Paszke, Gross, Massa, Lerer, Bradbury, Chanan,
  Killeen, Lin, Gimelshein, Antiga, Desmaison, Kopf, Yang, DeVito, Raison,
  Tejani, Chilamkurthy, Steiner, Fang, Bai, and Chintala]{pytorch}
Adam Paszke, Sam Gross, Francisco Massa, Adam Lerer, James Bradbury, Gregory
  Chanan, Trevor Killeen, Zeming Lin, Natalia Gimelshein, Luca Antiga, Alban
  Desmaison, Andreas Kopf, Edward Yang, Zachary DeVito, Martin Raison, Alykhan
  Tejani, Sasank Chilamkurthy, Benoit Steiner, Lu~Fang, Junjie Bai, and Soumith
  Chintala.
\newblock Pytorch: An imperative style, high-performance deep learning library.
\newblock In H.~Wallach, H.~Larochelle, A.~Beygelzimer, F.~d\textquotesingle
  Alch\'{e}-Buc, E.~Fox, and R.~Garnett, editors, \emph{Advances in Neural
  Information Processing Systems 32}, pages 8024--8035. Curran Associates,
  Inc., 2019.

\bibitem[Pathak et~al.(2017)Pathak, Agrawal, Efros, and Darrell]{pathak2017}
Deepak Pathak, Pulkit Agrawal, Alexei~A. Efros, and Trevor Darrell.
\newblock Curiosity-driven exploration by self-supervised prediction, 2017.

\bibitem[Pearl et~al.(2016)Pearl, Glymour, and Jewell]{pearl2016}
J~Pearl, M~Glymour, and N~P Jewell.
\newblock \emph{{Causal Inference in Statistics: A Primer}}.
\newblock Wiley, 2016.
\newblock ISBN 9781119186847.

\bibitem[Pearl(2009)]{Pearl2009}
Judea Pearl.
\newblock \emph{Causality: Models, Reasoning and Inference}.
\newblock Cambridge University Press, USA, 2nd edition, 2009.
\newblock ISBN 052189560X.

\bibitem[Singh et~al.(2005)Singh, Barto, and Chentanez]{Singh2005}
Satinder Singh, Andrew~G Barto, and Nuttapong Chentanez.
\newblock {Intrinsically Motivated Reinforcement Learning}.
\newblock \emph{IEEE Transactions on Autonomous Mental Development}, 2\penalty0
  (2):\penalty0 70--82, 2005.
\newblock ISSN 19430604.
\newblock \doi{10.1109/TAMD.2010.2051031}.

\bibitem[Song et~al.(2019)Song, Wang, Lukasiewicz, Xu, Zhang, Wojcicki, and
  Xu]{song2019megareward}
Yuhang Song, Jianyi Wang, Thomas Lukasiewicz, Zhenghua Xu, Shangtong Zhang,
  Andrzej Wojcicki, and Mai Xu.
\newblock Mega-reward: Achieving human-level play without extrinsic rewards,
  2019.

\bibitem[Sutton and Barto(2018)]{Sutton2018}
{Richard S.} Sutton and Andrew~G. Barto.
\newblock \emph{{Reinforcement Learning - An Introduction}}.
\newblock 2018.
\newblock ISBN 9780262039246.

\bibitem[Vinyals et~al.(2019)Vinyals, Babuschkin, Czarnecki, Mathieu, Dudzik,
  Chung, Choi, Powell, Ewalds, Georgiev, Oh, Horgan, Kroiss, Danihelka, Huang,
  Sifre, Cai, Agapiou, Jaderberg, and Silver]{AlphaStar2019}
Oriol Vinyals, Igor Babuschkin, Wojciech Czarnecki, Michaël Mathieu, Andrew
  Dudzik, Junyoung Chung, David Choi, Richard Powell, Timo Ewalds, Petko
  Georgiev, Junhyuk Oh, Dan Horgan, Manuel Kroiss, Ivo Danihelka, Aja Huang,
  Laurent Sifre, Trevor Cai, John Agapiou, Max Jaderberg, and David Silver.
\newblock Grandmaster level in starcraft ii using multi-agent reinforcement
  learning.
\newblock \emph{Nature}, 575, 11 2019.
\newblock \doi{10.1038/s41586-019-1724-z}.

\bibitem[Wang et~al.(2015)Wang, de~Freitas, and Lanctot]{Wang2015}
Ziyu Wang, Nando de~Freitas, and Marc Lanctot.
\newblock Dueling network architectures for deep reinforcement learning.
\newblock \emph{CoRR}, abs/1511.06581, 2015.

\end{thebibliography}

\newpage
\appendix
\appendixpage

\section{Additional experiments}
\label{app:experiments}

\subsection{Effect of Alpha on CEN}
\label{app:ablation}

Here we study how $\alpha$ affects the loss in Eq. \ref{eq:loss}. This experiment uses Montezuma's Revenge and the same setup as in Experiment \ref{sec:experiments_pixel}. We analyze the effect of alpha on CEN using values ranging from 0.01 to 20.
\begin{figure}[H]
    \centering
    \includegraphics[width=0.7\textwidth]{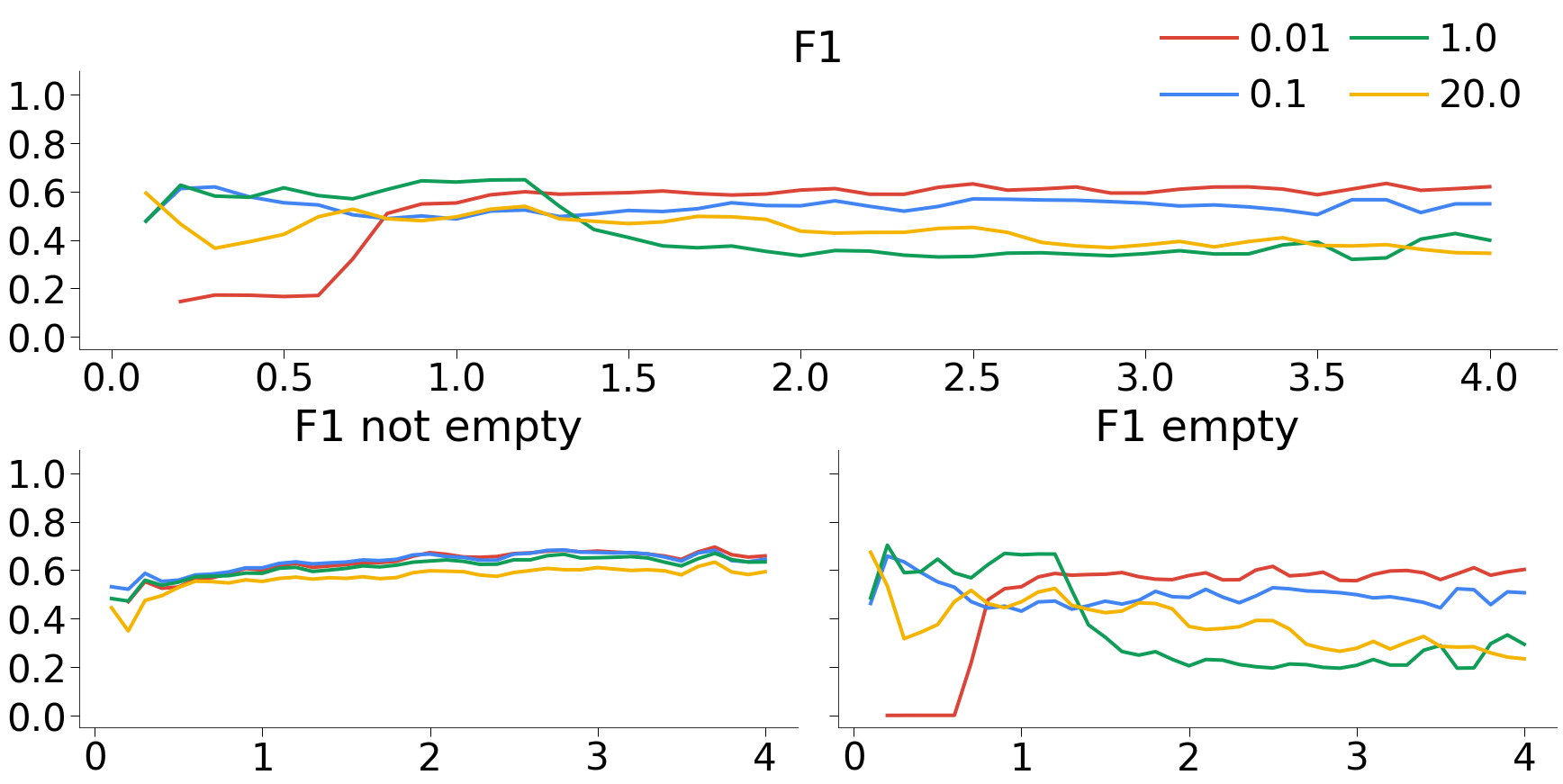}
    \caption{Ablation study on the effect of $\alpha$ in Eq. \ref{eq:loss} on CEN in MZR.}
    \label{fig:alpha}
\end{figure}

As can be seen in Fig. \ref{fig:alpha}, the different alphas do not impact the performance when the ground truth or controlled effects are not empty. On the other hand, the performance degrades the higher the alpha.
We hypothesize that this behavior happens when the normal branch is forced to model controlled effects too strongly and the controlled branch needs to counter those bad predictions. In this case, the controlled branch will produce wrong masks, especially for empty ground truth.
\begin{figure}[H]
    \includegraphics[width=\textwidth]{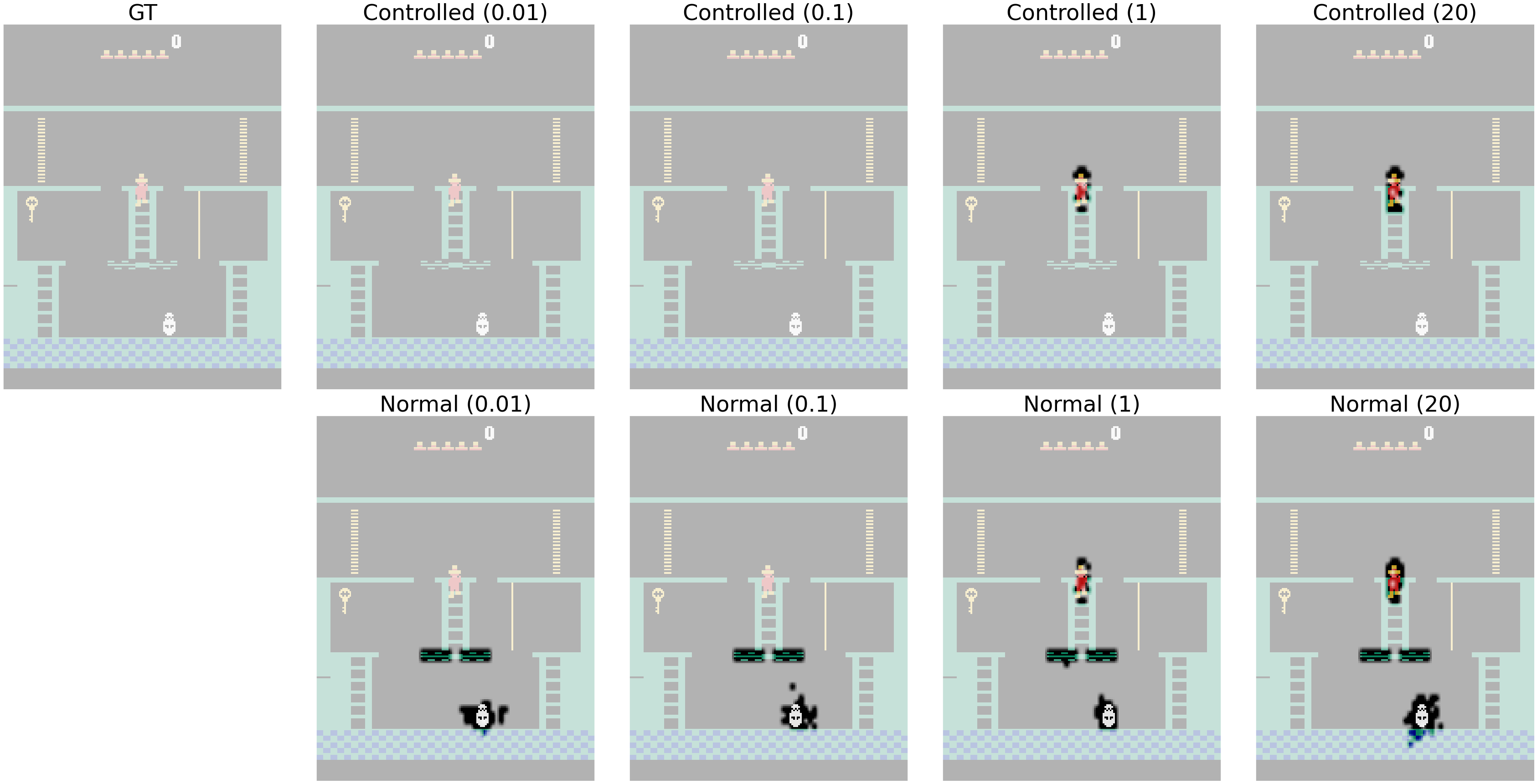}
    \caption{Examples of masks for $\alpha=0.01$ and $\alpha=20$ in MZR. Masks generated by the controlled and normal branches for each $\alpha$ (in parenthesis) are highlighted (darker).}
    \label{fig:ablation_masks}
\end{figure}

\subsection{Masks}
\label{app:masks}
Here we provide masks for both CEN and ADM. Mask are based on effects and extend over two frames, for visualization purposes here we only show the next observation with the full mask.
It can be seen that CEN fails for a few pixels in MZR, which may explain the low F1 score for empty GT.
\begin{figure}[H]
    \centering
    \begin{subfigure}{0.85\textwidth}
        \includegraphics[width=\textwidth]{images/mask_mzr_1.png}
    \end{subfigure}
    \begin{subfigure}{0.85\textwidth}
        \includegraphics[width=\textwidth]{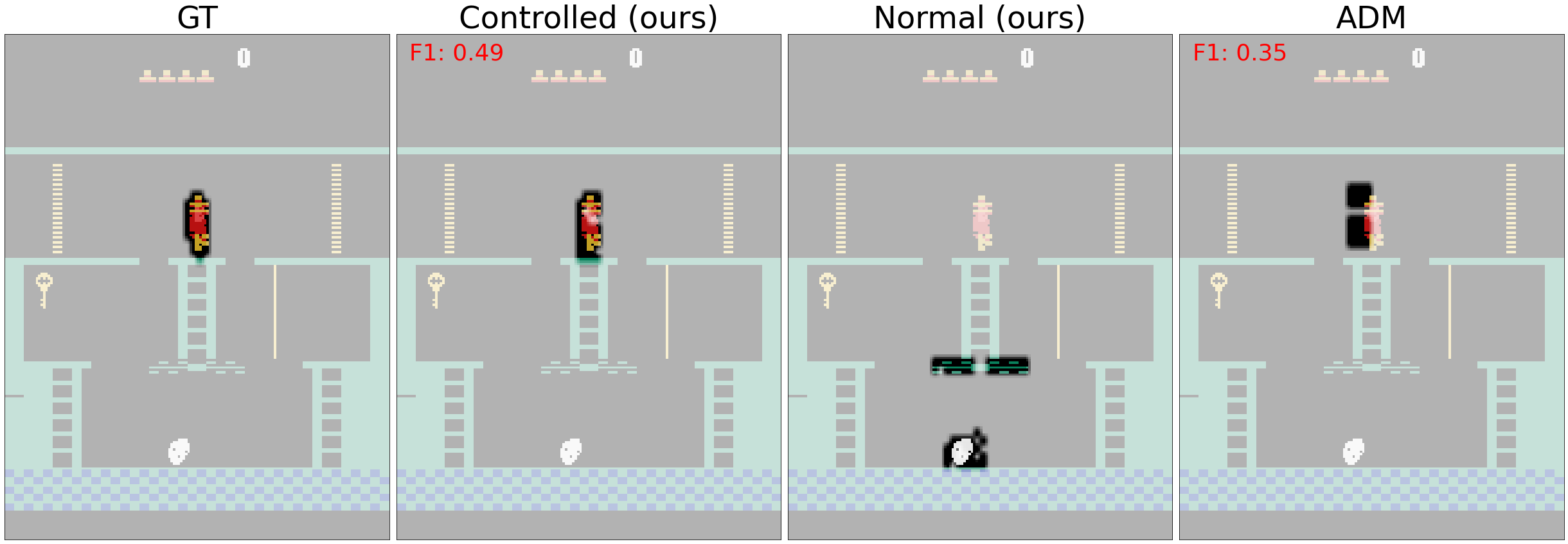}
    \end{subfigure}
    \begin{subfigure}{0.85\textwidth}
        \includegraphics[width=\textwidth]{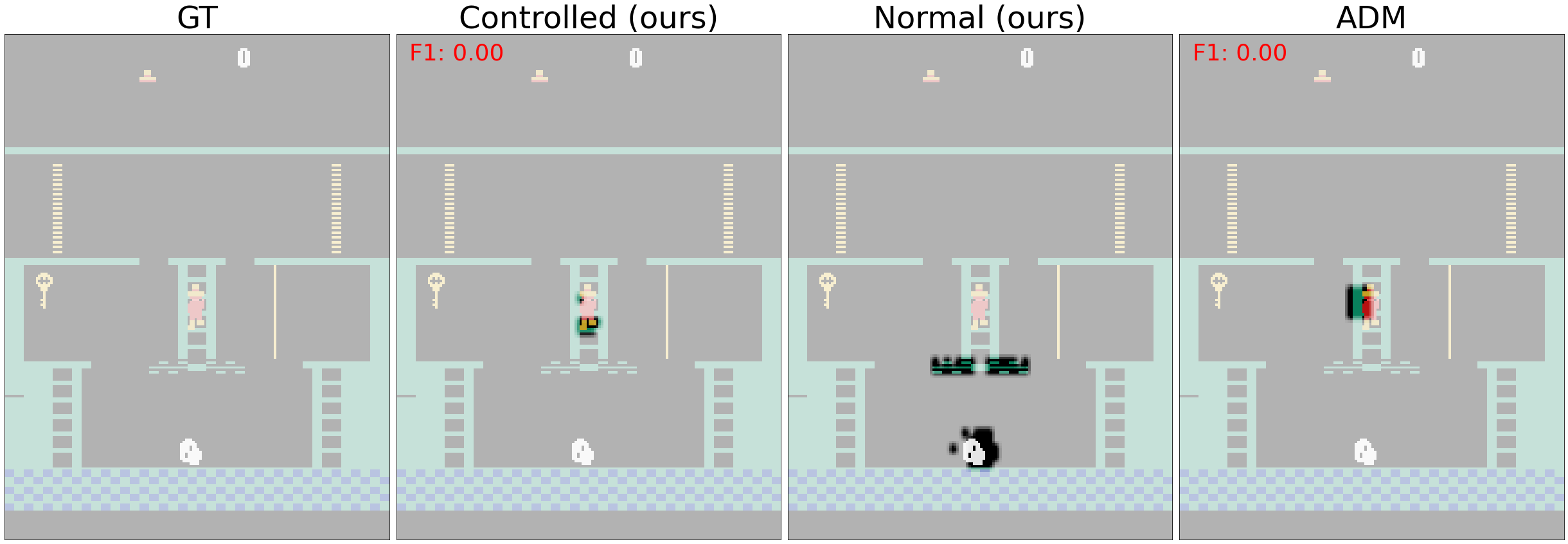}
    \end{subfigure}
    \begin{subfigure}{0.85\textwidth}
        \includegraphics[width=\textwidth]{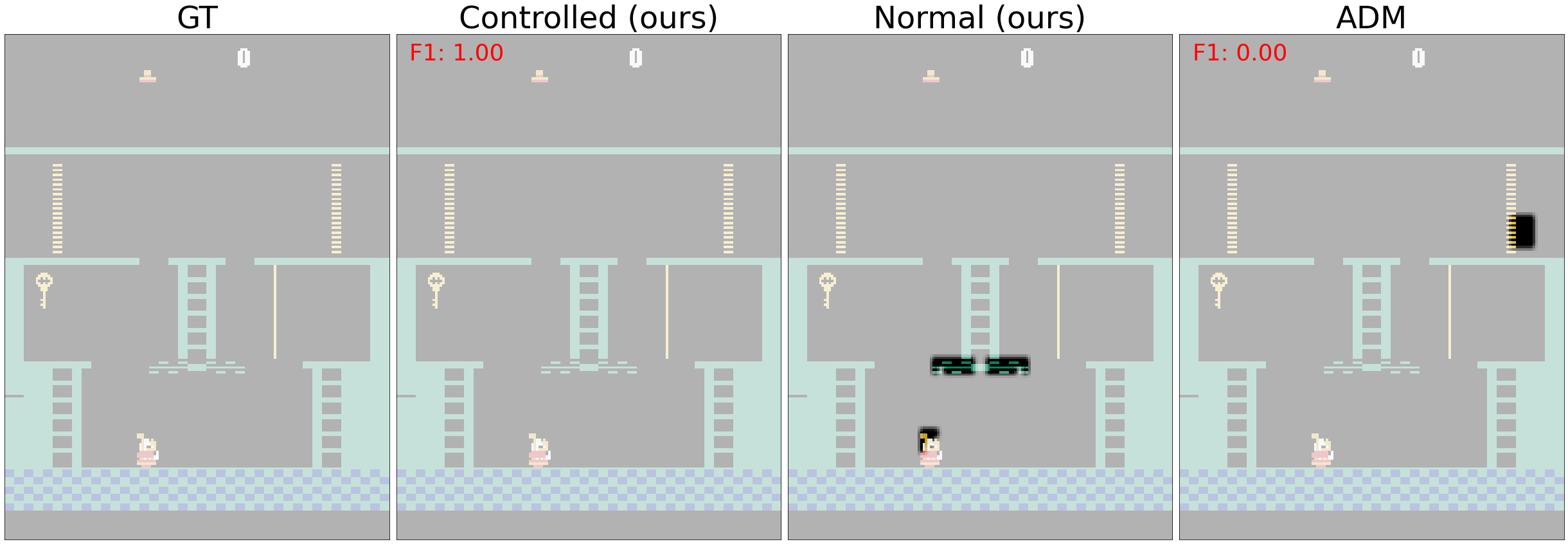}
    \end{subfigure}
    \caption{Examples of success and failure cases for CEN and ADM in MZR.}
    \label{fig:masks_mzr}
\end{figure}

\begin{figure}[H]
    \centering
    \begin{subfigure}{0.85\textwidth}
        \includegraphics[width=\textwidth]{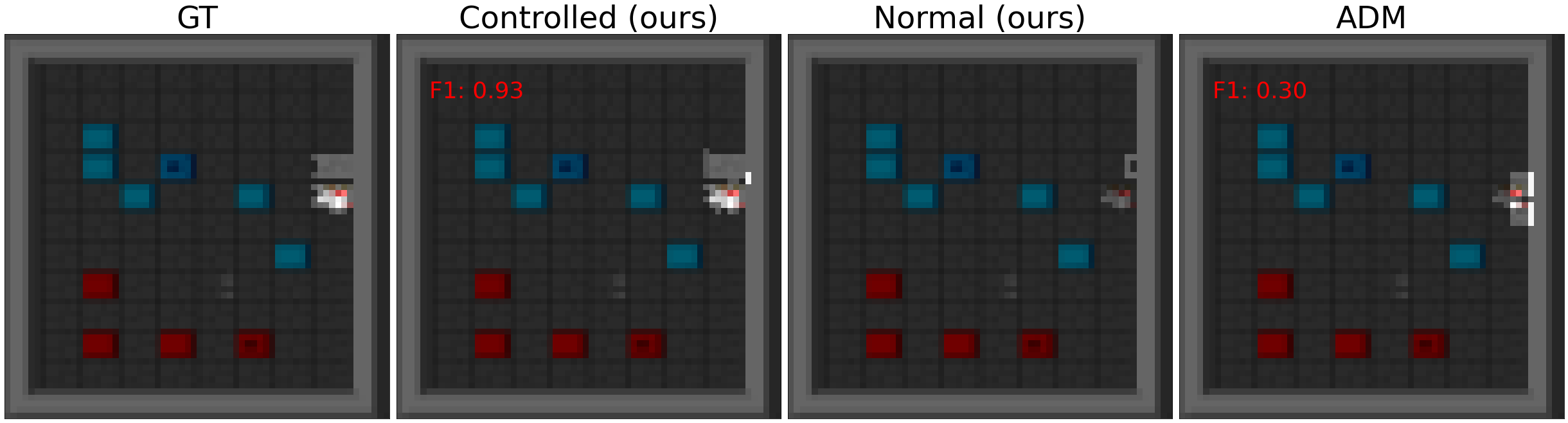}
    \end{subfigure}
    \begin{subfigure}{0.85\textwidth}
        \includegraphics[width=\textwidth]{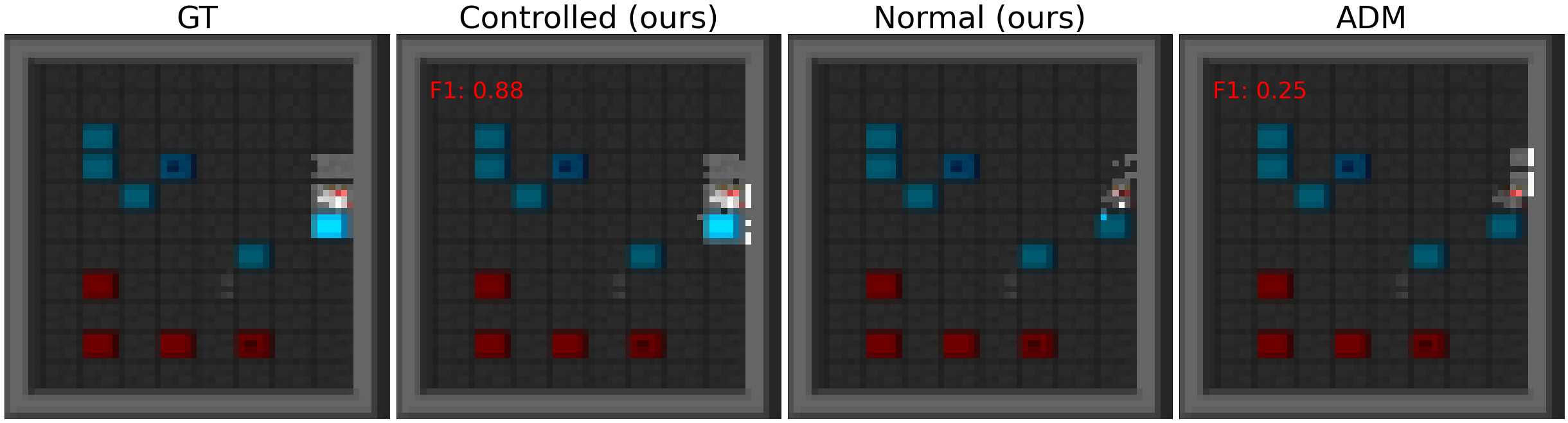}
    \end{subfigure}
    \caption{Examples of masks for Clusters.}
    \label{fig:masks_clusters}
\end{figure}

\begin{figure}[H]
    \centering
    \begin{subfigure}{0.85\textwidth}
        \includegraphics[width=\textwidth]{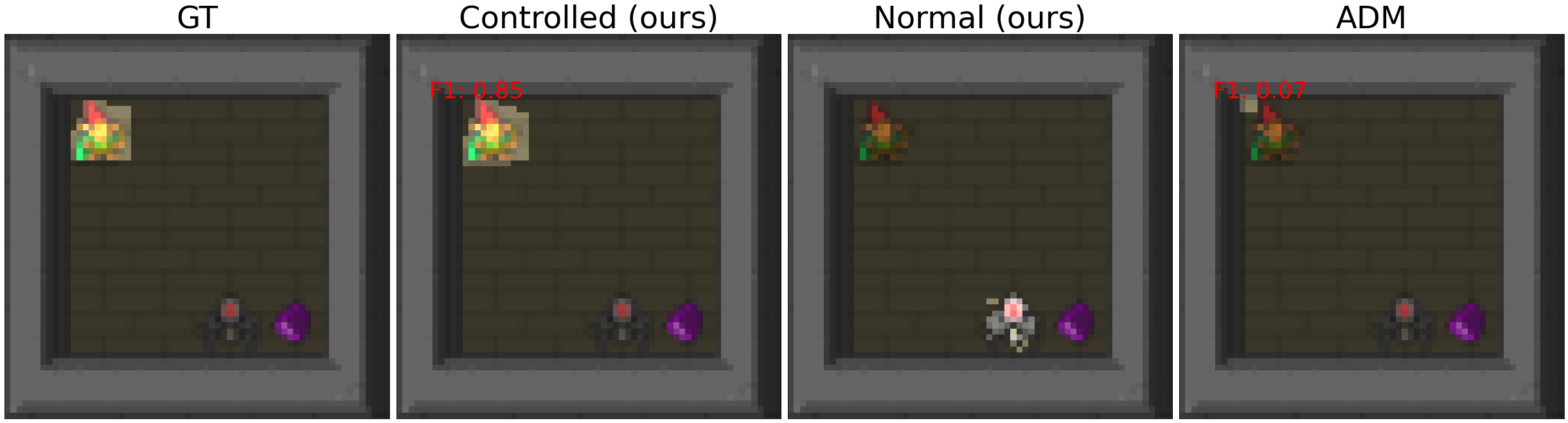}
    \end{subfigure}
    \begin{subfigure}{0.85\textwidth}
        \includegraphics[width=\textwidth]{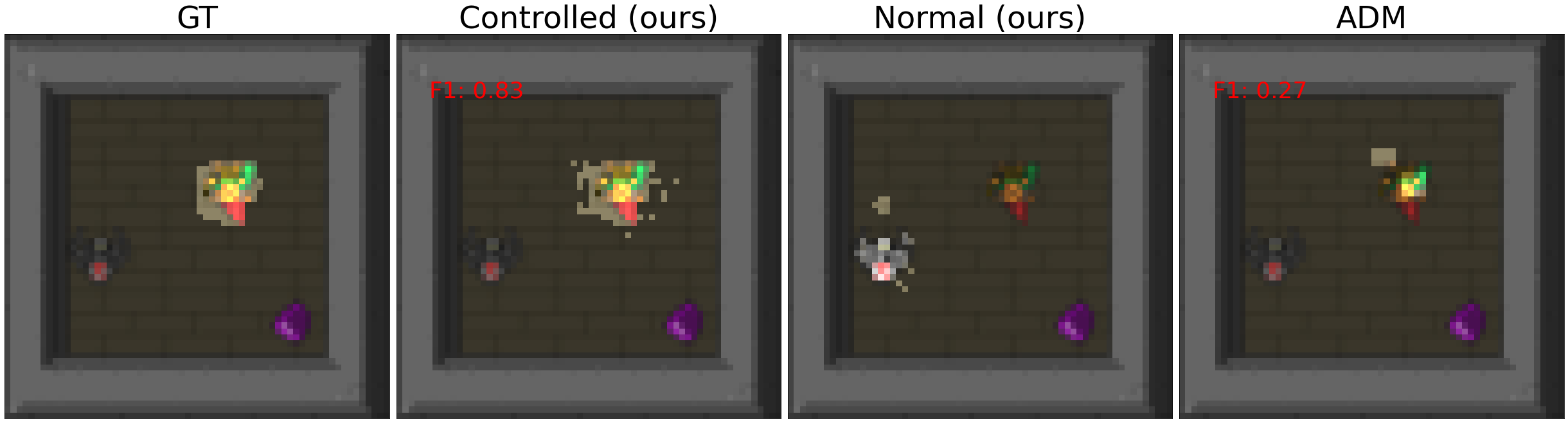}
    \end{subfigure}
    \begin{subfigure}{0.85\textwidth}
        \includegraphics[width=\textwidth]{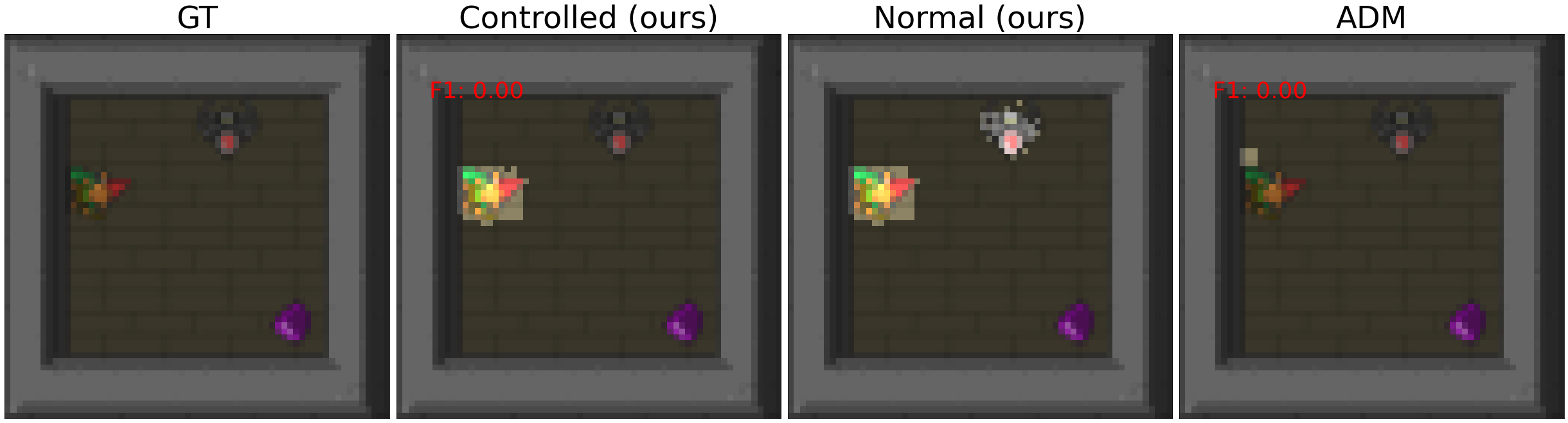}
    \end{subfigure}
    \caption{Examples of masks for Spiders.}
    \label{fig:masks_spiders}
\end{figure}

\begin{figure}[H]
    \centering
    \begin{subfigure}{0.85\textwidth}
        \includegraphics[width=\textwidth]{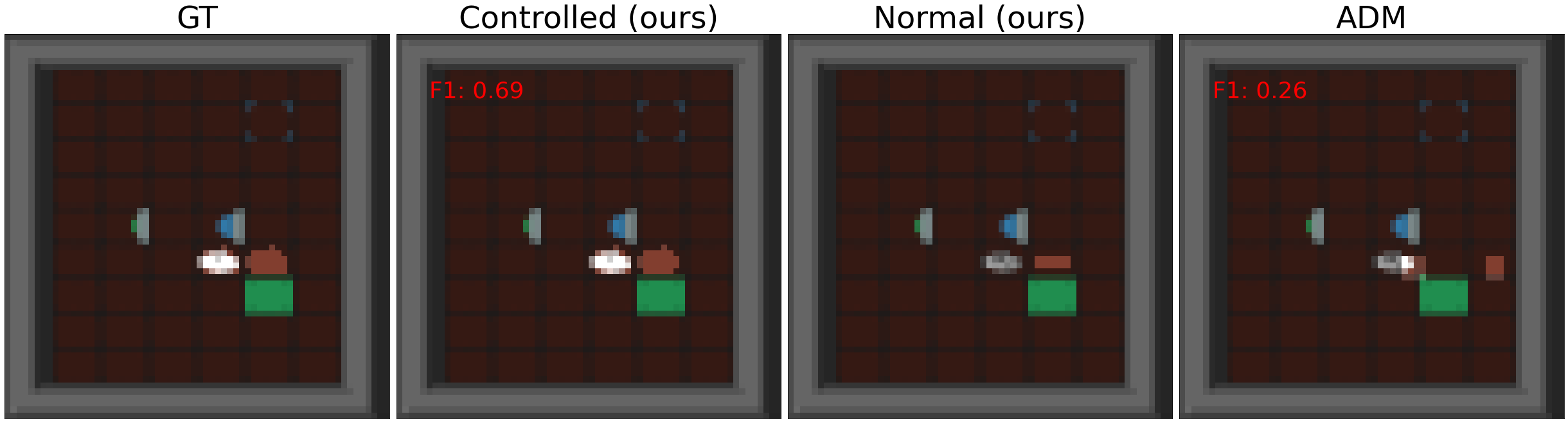}
    \end{subfigure}
    \begin{subfigure}{0.85\textwidth}
        \includegraphics[width=\textwidth]{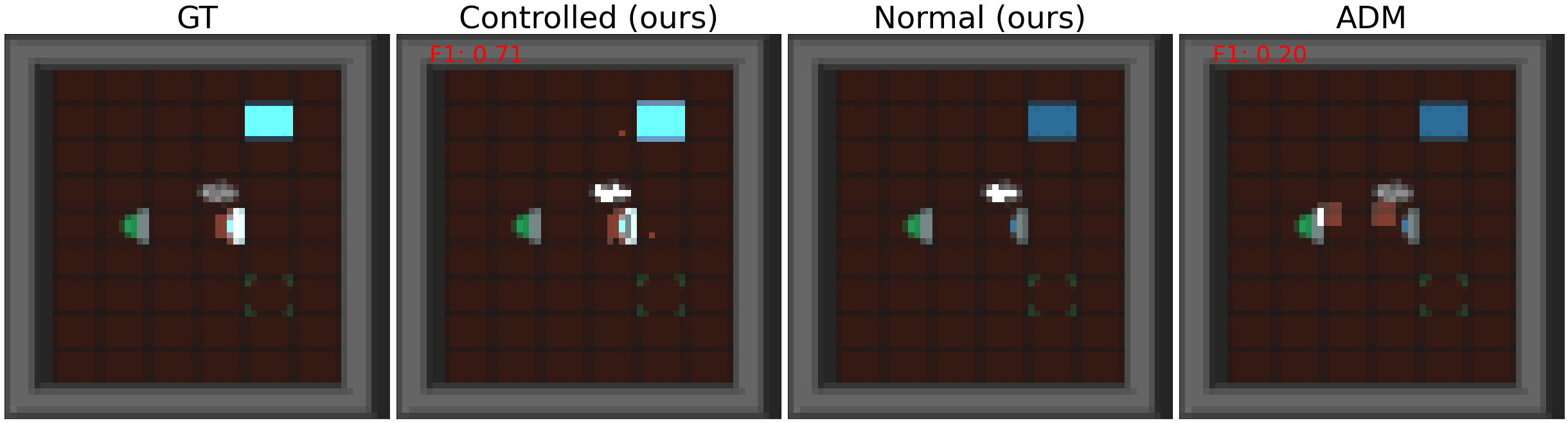}
    \end{subfigure}
    \caption{Examples of masks for Lights.}
    \label{fig:masks_lights}
\end{figure}

\section{Environments}
\label{app:envs}

\subsection{Griddly}
\label{app:griddly}
Griddly \citep{bamford2020griddly} is a highly optimized grid-world based suite of environments.
Environments used in this work based on Griddly generate $64\times64$ pixel observations, although the size of the grid-world may vary. Griddly supports multiple rendering formats, this work uses the 2D rendering of sprites.
\\\\
\textbf{Spiders:} is a 6x6 arena where a Gnome (the agent) has to grab a Gem without being killed by a Spider.
The agent dies if it collides with the spider.
In this environment, the agent can move \textit{left}, \textit{right}, \textit{up}, \textit{down} or \textit{stay}.
The spider takes an action randomly from the following: rotate left, rotate right or move forward.
This environment's controlled entities are: Agent.
\\\\
\textbf{Clusters:} is a 13x10 arena where a Knight has to move boxes of the same color to their corresponding colored-block without touching the spikes.
There are two different colors, blue and red. The agent is rewarded with +1 whenever a box is pushed towards a similar colored block.
The agent dies if it collides with spikes or if a box is destroyed by spikes.
The agent can move \textit{left}, \textit{right}, \textit{up}, \textit{down} or \textit{stay}.
This environment's controlled entities are: Agent and Boxes.\\
For RL experiments, we made the environment more sparse by removing all intermediate reward and only rewarded the agent after all the boxes of the same color are pushed to blocks. Since the agent can not get any reward whenever a box is stuck to a wall. We removed boxes that touches the wall and punished the agent with -0.01. We then scaled the reward of solving a color with the number of boxes pushed to the block. This modifications encourage the agent to solve the environment by pushing the maximum number of boxes into blocks while preventing it from getting deprived from reward by accidentally pushing boxes to walls.
\\\\
\textbf{Lights:} is a 11x8 arena where a Ghost (the agent) has to turn all the lights on by pressing each button.
Buttons and lights are colored either blue or green. Pressing a button of one color turns the light of the same color on.
The agent can move \textit{left}, \textit{right}, \textit{up}, \textit{down} or \textit{stay}.
This environment's controlled entities are: Agent, Buttons and Lights.
\\\\
\textbf{Empty:} this environment is a copy of the clusters environment where all the boxes, blocks, spikes and rewards were removed.
\\\\
\textbf{Ground truth:} Griddly provides access to each entity's state (x, y, light is on/off, etc); a binary mask is produced for each controlled entity with any of its attributes changed when transitioning between two time steps. x and y coordinates are projected into pixel space and a bounding box is generated using the size of that entity. Note that the coordinates may be different between steps thus the generated mask may enclose multiple locations. The resulting masks for each entity are combined into a single mask $m$ by taking the maximum value among them. Since we want to know what pixels were actually controlled, the final ground truth mask is produced as: $m \cdot e^p_a$.

\subsection{Atari Montezuma's Revenge}
The ALE \citep{bellemare2012ale} provides access to Atari 2600 games to learning methods like RL. As it has been a popular choice by methods using inverse models, in this work we use the game Montezuma's Revenge to evaluate CEN. This environment provides uncontrolled as well as controlled effects with more complex entities.
The environment typically generates observations of 210x160 pixels which we downscale to 64x64 pixels. Additionally the action space is of size 10.
\\\\
\textbf{Ground truth:} in contrast to Griddly, we can actually compute counterfactual worlds by saving and loading the state of the game (RAM) multiple times when taking different actions. For this, we directly compute Eq. 5 using ALE's special calls \textit{cloneSystemState} and \textit{restoreSystemState}.
More precisely, we compute every possible perceived effect reachable from the current state and build a normal effect using the mode over all possible effects. Then, we compare the perceived effect for the agent's chosen action against the normal effect. 
The ground truth mask will have 1s where these two effects are different.

\section{Training}

\subsection{Architecture}
\textbf{Encoder:} is composed of two 2D convolutional layers with 4x4 kernels, stride 2 and padding of 1. Additionally, we have 2 residual blocks each with two 2D convolutional layers with stride 1 and padding of 1. The first layer has a kernel of 3x3 and the second layer of 1x1. ReLU is used as activation function throughout the network; BatchNorm is used between each layer; and 64 channels on every convolutional layer. We project the resulting maps into a flatten vector of size 32 using a linear layer with ReLU activation function.
\\\\
\textbf{Decoder:} this module is composed of six 2D transposed convolutional layers all having 4x4 kernels, stride 2 and padding of 1. Each layer uses ReLU as activation function but the output layer which uses Tanh activation. Every layer uses 64 channels with the exception of the last layer which outputs a 1 channel prediction of the perceived effects. Parameters are shared among the controlled and normal branch decoders.
\\\\
\textbf{Controlled and normal modules:} both modules are composed of three linear layers with 32 hidden units, each with a ReLU as activation function. The input to the controlled branch are the encoded observation and an embedding of size 8 of the chosen action.
\\\\
\textbf{PPO Agent:} uses an encoder consisting of 3 convolutional layers with (channels, padding, strides) equal to (32, 8, 4), (64, 4, 2), (64, 3, 1) respectively. The encoder is followed by two linear layers of sizes 512 and number of actions respectively, to transform the feature map to the environment's number of actions. 

\subsection{Mask generation}
CEN's controlled masks are generated using the encoder, controlled branch and decoder. The predicted controlled effect is binarized using a threshold as $m_\text{CEN} = (-T < \hat{e^c}) | (\hat{e^c} > T)$.
In the case of ADM, its attention mask is thresholded in the same way and resized to the size of the effect.

\subsection{Hyperparameters}
\label{app:hyperparameters}

\begin{table}[H]
\begin{center}
\begin{tabular}{ c| c| c }
 Name & Value & Sweep \\ 
 \hline
 hidden size & 32 & [16, 32, 64, 128] \\
 latent size & 128 & [16, 32, 64, 128, 256] \\
 channels & 64 & [16, 32, 64, 128] \\
 learning rate & 0.0001 & [0.0001, 0.0005, 0.001, 0.005]\\
 $\alpha$ & 0.01 & [0.001, 0.01, 0.1, 1, 5, 10, 20, 30, 50] \\
 $T$ & 0.01 & - \\
\end{tabular}
\label{tab:hyperparameters_cen}
\caption{CEN hyperparameter sweeps and final values used.}
\end{center}
\end{table}

\begin{table}[H]
\begin{center}
\begin{tabular}{ c| c| c }
 Name & Value & Sweep \\ 
 \hline
 entropy & 0.05 & [0.01, 0.05, 0.1, 0.5, 1, 5] \\
 hidden size & 64 & [16, 32, 64, 128] \\
 attention size & 128 & [32, 64, 128, 256] \\
 learning rate & 0.0001 & [0.0001, 0.0005, 0.001, 0.005]\\
 $T$ & 0.01 & - \\
\end{tabular}
\label{tab:hyperparameters_adm}
\caption{ADM hyperparameter sweeps and final values used.}
\end{center}
\end{table}

\begin{table}[H]
\begin{center}
\begin{tabular}{ c| c| c }
 Name & Value & Sweep \\ 
 \hline
 encoder channels & 32 & [32, 64] \\
 encoder hidden & 32 & [16, 32, 64, 128] \\
 latent size & 128 & [64, 128, 256] \\
 learning rate & 0.0001 & [0.0001, 0.0005, 0.001, 0.005]\\
\end{tabular}
\label{tab:hyperparameters_inverse}
\caption{Inverse model hyperparameter sweeps and final values used.}
\end{center}
\end{table}

\begin{table}[H]
\begin{center}
\begin{tabular}{ c| c| c }
 Name & Value & Sweep \\ 
 \hline
 batch size & 512 & [64, 128, 512, 1024] \\
 latent size & 32 & [8, 16, 32] \\
 CEN encoder output size & 128 & [16, 32, 64, 128] \\
 learning rate & 0.0005 & [0.00005, 0.0001, 0.0002, 0.0005, 0.001]\\
 IR Beta & 0.001 & [0.0001, 0.001, 0.002, 0.005, 0.01, 0.1] \\
 Rollout size & 2048 & [1024, 2048, 4096] \\
 discount & 0.95 & - \\
 epochs & 10 & - \\
\end{tabular}
\label{tab:hyperparameters_ppo}
\caption{PPO hyperparameter sweeps and final values used.}
\end{center}
\end{table}

\end{document}